\documentclass[11pt]{article}

\usepackage[preprint]{acl}

\usepackage{times}
\usepackage{latexsym}

\usepackage[T1]{fontenc}
\usepackage[utf8]{inputenc}

\usepackage{microtype}

\usepackage{inconsolata}

\usepackage{graphicx}

\usepackage{amsmath}
\usepackage{booktabs}
\usepackage{pdfpages}
\usepackage{graphicx}
\usepackage{caption}
\usepackage{multirow}
\usepackage{cleveref}
\usepackage{xspace}
\usepackage{diagbox}
\usepackage{hyperref}
\usepackage{tabularx}
\usepackage{hypcap}
\usepackage{enumitem}
\usepackage{supertabular}
\usepackage{longtable}
\usepackage{algorithm}
\usepackage{algpseudocode}
\usepackage{todonotes}
\usepackage{fnpct}

\title{COMMUNITYNOTES:\\ A Dataset for Exploring the Helpfulness of Fact-Checking Explanations}

\author{
 \textbf{Rui Xing\textsuperscript{1,2}} \ 
 \textbf{Preslav Nakov\textsuperscript{2}} \ 
 \textbf{Timothy Baldwin\textsuperscript{1,2}} \
 \textbf{Jey Han Lau\textsuperscript{1}} \\  
 \textsuperscript{1}The University of Melbourne,
 \textsuperscript{2}MBZUAI\\
\texttt{ruixing@student.unimelb.edu.au,preslav.nakov@mbzuai.ac.ae} \\ 
\texttt{tb@ldwin.net, jeyhan.lau@gmail.com} \\
}

\newcommand{\model}[1]{\text{#1}\xspace}
\newcommand{\dataset}[1]{\texttt{#1}\xspace}
\newcommand{\framework}[1]{\textsc{#1}\xspace}
\newcommand{\website}[1]{\textit{#1}\xspace}

\newcommand{\ourdataset}{\dataset{COMMUNITYNOTES}}
\newcommand{\climatefever}{\dataset{CLIMATE-FEVER}}
\newcommand{\promptagent}{\framework{PromptAgent}}
\newcommand{\noteswebsite}{\website{Community Notes}}

\newcommand{\bert}{\model{BERT}}
\newcommand{\bertbase}{\model{BERT-base}}
\newcommand{\bertlarge}{\model{BERT-large}}
\newcommand{\roberta}{\model{RoBERTa}}
\newcommand{\robertabase}{\model{RoBERTa-base}}
\newcommand{\robertalarge}{\model{RoBERTa-large}}
\newcommand{\modernbert}{\model{ModernBERT}}
\newcommand{\modernbertbase}{\model{ModernBERT-base}}
\newcommand{\modernbertlarge}{\model{ModernBERT-large}}
\newcommand{\deberta}{\model{DeBERTa}}
\newcommand{\debertabase}{\model{DeBERTa-base}}
\newcommand{\debertalarge}{\model{DeBERTa-large}}
\newcommand{\mbert}{\model{mBERT}}
\newcommand{\xlm}{\model{XLM-RoBERTa}}
\newcommand{\xlmbase}{\model{XLM-RoBERTa-base}}
\newcommand{\xlmlarge}{\model{XLM-RoBERTa-large}}

\newcommand{\llama}{\model{Llama3.1-8B}}
\newcommand{\mistral}{\model{Mistral-7B}}

\newcommand{\gptfouro}{\model{GPT-4o}}
\newcommand{\z}{\phantom{0}}

\begin{document}
\maketitle
\begin{abstract}
Fact-checking on major platforms, such as X, Meta, and TikTok, is shifting from expert-driven verification to a community-based setup, where users contribute explanatory notes to clarify why a post might be misleading. An important challenge here is determining whether an explanation is helpful for understanding real-world claims and the reasons why, which remains largely underexplored in prior research. In practice, most community notes remain unpublished due to slow community annotation, and the reasons for helpfulness lack clear definitions. To bridge these gaps, we introduce the task of predicting both the helpfulness of explanatory notes and the reason for this. We present \ourdataset, a large-scale multilingual dataset of 104k posts with user-provided notes and helpfulness labels. We further propose a framework that automatically generates and improves reason definitions via automatic prompt optimization, and integrate them into prediction. Our experiments show that the optimized definitions can improve both helpfulness and reason prediction. Finally, we show that the helpfulness information is beneficial for existing fact-checking systems. The code and the data are available at \url{https://github.com/ruixing76/Helpfulness-FCExp}.
\end{abstract}

\begin{figure}[t]
    \centering
    \includegraphics[width=1\linewidth]{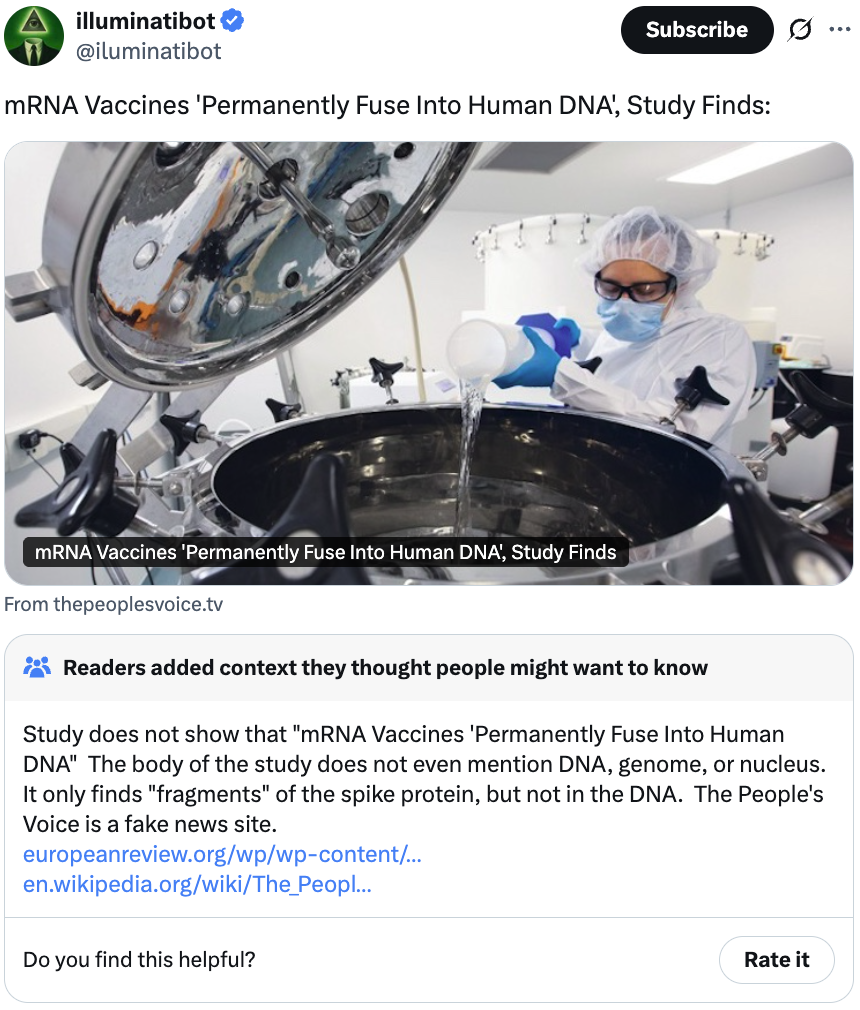}
    \caption{An example of X Community Notes. The potential misleading tweet states, ``mRNA Vaccines Permanently Fuse Into Human DNA''. The user-generated note appears under \emph{Readers added context} and provides an explanation for the tweet.}
    \label{fig:example}
\end{figure}
% example link: https://x.com/iluminatibot/status/1775074443833770213

\section{Introduction}
Beyond simple verification, fact-checking systems should provide explanations that clarify why a claim is misleading~\citep{atanasova-etal-2020-generating-fact, xing-etal-2022-automatic, russo-etal-2023-benchmarking,explainablefc}. Explanations not only strengthen user trust in fact-checking decisions, but also help mitigate common pitfalls of fact-checking, e.g., backfire effect, where exposure to corrections can paradoxically reinforce false beliefs rather than hinder them~\citep{stephan-misinformation-2012,guo-etal-2022-survey, xing-etal-2025-evaluating}.

Typically, fact-checking explanations are produced by expert organizations. Yet the ecosystem is currently shifting, and platforms are increasingly experimenting with community-based fact-checking, where contributors collaboratively provide explanatory notes on claims~\citep{communitymoderationnewepistemology}. This crowd-sourcing process distributes the labor of verification and offers diverse perspectives from the community. A representative example is \emph{Community Notes}\footnote{\url{https://communitynotes.x.com/guide/}} on X (formerly Twitter), which allows users to attach notes to explain potentially misleading posts. As illustrated in \Cref{fig:example}, the user-generated note appears under \emph{Readers added context} and provides an explanation for the tweet. 

Critically, for a note to become visible or formally published, it must receive a sufficient number of ratings from a diverse set of users~\citep{supernotes}. A key challenge in this process is determining whether an explanation is helpful for understanding real-world claims, as well as identifying why it is helpful, a question that remains largely underexplored in prior research on fact-checking explanations~\citep{atanasova-etal-2020-generating-fact, guo-etal-2022-survey, russo-etal-2023-benchmarking, eldifrawi-etal-2024-automated}. In practice, this lack of clarity makes it difficult for contributors to write effective notes and for raters to assess them consistently.

This process is slow on \noteswebsite: notes can take hours or even days to appear, and the vast majority (over 90\%) never become visible, despite contributor efforts~\citep{supernotes}. Moreover, although the platform provides reasons for judging notes as helpful or unhelpful, these reasons lack clear definitions, leaving both contributors and raters unclear about the underlying evaluation criteria~\citep{supernotes,communitymoderationnewepistemology}. This opacity hinders the efficiency, scalability, and explainability of the crowd-sourced fact-checking. Motivated by these challenges, we explore the idea of simplifying and making the helpfulness annotation process more explicit. Our contributions of this paper are as follows:

\begin{itemize}
    \item We introduce the task of predicting the helpfulness of an explanatory note and the reasons underlying this judgment, given the note and the original X post.
    \item We present \ourdataset, a large-scale dataset for explanation helpfulness, containing over 104K potentially misleading posts with corresponding user-provided notes.
    \item We propose a framework that automatically generates and optimizes reason label definitions via prompt optimization, and integrates these definitions into helpfulness and reason predictors to improve their performance.
    \item We apply explanation helpfulness prediction to the evidence sufficiency task~\citep{atanasova-etal-2022-fact} and further incorporate it as an auxiliary signal into automated fact-checking on \climatefever~\citep{diggelmann2020climatefever}, demonstrating its potential to enhance current existing automated fact-checking systems.
\end{itemize}

\section{Related Work}

\paragraph{Community-Based Fact-Checking} 
The proliferation of misinformation has become a growing societal concern, posing significant risks to areas such as public health, science, and politics~\citep{borenstein-etal-2025-community}. Misleading claims can spread rapidly online, shaping public opinion and undermining trust in institutions. To address this challenge, fact-checking has emerged as a central strategy for mitigating the spread of false or misleading information, by evaluating claims and providing corrective context to users~\citep{guo-etal-2022-survey,afc_nakov_2021,xie-etal-2025-fire}. 

Traditionally, fact-checking has been carried out by professional third-party organizations that assess claim veracity through expert analysis. However, as a result, recent years have seen a paradigm shift toward more decentralized, community-based approaches~\citep{renault_collaboratively_2024,communitymoderationnewepistemology}. Major social media platforms have increasingly adopted this paradigm. X's \noteswebsite is the most prominent example, allowing their users to collaboratively annotate and assess online content by attaching contextual notes to potentially misleading posts. Following this trend, Meta ended its partnerships with professional fact-checkers\footnote{\href{https://www.poynter.org/fact-checking/2025/meta-ends-fact-checking-community-notes-facebook}{Meta is ending its third-party fact-checking partnership with US partners. Here’s how that program works.}} in favor of a similar community-driven system, while TikTok has also introduced a comparable feature.\footnote{\href{https://newsroom.tiktok.com/en-us/footnotes}{Testing a New Feature to Enhance Content on TikTok.}}

Empirical evidence suggests that community-based fact-checking can be highly effective in countering misinformation at scale. Studies have shown that the presence of a Community Note can reduce the rate of retweets by nearly half~\citep{renault_collaboratively_2024}, while displaying such notes on misleading posts has been found to decrease their spread by an average of 61\%~\citep{chuai_community_2024}. Moreover, authors of posts flagged by community notes are up to 80\% more likely to delete their content, and such posts generally exhibit markedly lower virality~\citep{diffusionnotes,renault_collaboratively_2024}. These findings highlight the potential of community-based explanations in debunking online misinformation.

Despite these benefits, \emph{Community Notes} often require a long time to accumulate sufficient contributor ratings to determine helpfulness. In addition, the absence of reason definitions further undermines the trustworthiness of the mechanism.

\paragraph{Label Definitions} Recent work has shown that providing models with explicit definitions of labels can lead to substantial improvements in classification. \citet{khatuya-etal-2025-label} used predefined label descriptions and trained the model to generate these descriptions based on the input text, achieving sizable improvements in multi-label classification. Similarly, \citet{gao-etal-2023-benefits} demonstrated that curating a small dictionary of terms for label description can improve zero-shot accuracy by 17-19\% absolute, while also increasing the robustness to prompts. \citet{peskine-etal-2023-definitions} showed improvements when using label definitions with GPT-3 zero-shot classification on the challenging task of fine-grained conspiracy theory detection.

\paragraph{Automatic Prompt Optimization} Prompt engineering has become essential to many Natural Language Processing (NLP) tasks; yet, manual trial-and-error approaches remain a barrier to effective use. Automatic Prompt Optimization (APO) methods address this challenge by systematically searching for high-performing prompts~\citep{ramnath2025systematicsurveyautomaticprompt}. These approaches explore the large instruction space using diverse strategies: \framework{PromptBreeder} evolves task-specific prompts across generations using optimized mutation strategies~\citep{promptbreeder}; \promptagent applies Monte Carlo Tree Search (MCTS) with feedback reflection to guide exploration~\citep{wang2024promptagent}; and \framework{SCULPT} represents prompts as trees, enabling targeted refinements while preserving overall structure—especially useful for long or complex prompts~\citep{kumar-etal-2025-sculpt}.

\section{Predicting Whether an Explanation is Helpful and Why}
\subsection{Task Definition}
Given a post $P$ that contains potentially misleading information and a note $N$ that provides an explanation about that post, the task is to predict two types of labels: a \textbf{helpfulness label} $L_{\mathit{helpful}}$, indicating whether the note is helpful in explaining the misleading nature of the post, and a \textbf{reason label} $L_{\mathit{reason}}$, specifying why the note is (un)helpful. The helpfulness prediction is formulated as a binary classification task with the label space $\{\mathit{Helpful}, \mathit{Unhelpful}\}$. The reason prediction is a multi-label classification task, where the predefined label set includes eight reasons for helpful notes and ten reasons for unhelpful ones (see Appendix \Cref{tab:original_prompt} for detailed reason labels).

\begin{table}[h]
\centering
\small
\resizebox{1\columnwidth}{!}{
\begin{tabular}{lcccc}
\toprule
\textbf{Language} & \textbf{Train} & \textbf{Dev} & \textbf{Test} & \textbf{Total} \\
\midrule
English & 40,994 & 5,858 & 11,717 & 58,569 \\
Other languages & 32,478 & 4,638 & 9,281 & 46,397 \\
\midrule
\textbf{Total} & 73,472 & 10,496 & 20,998 & 104,966 \\
\bottomrule
\end{tabular}
}
\caption{\ourdataset data split statistics for English and Other languages.}
\label{tab:data_split}
\end{table}

\begin{table}[h]
\centering
\small
\begin{tabular}{lccccc}
\toprule
\textbf{Text Type} & \textbf{Mean} & \textbf{Median} & \textbf{Min} & \textbf{Max} \\
\midrule
Post & 57.44 & 42 & 0 & 5,072 \\
Note  & 87.05 & 70 & 1 & 1,402 \\
\bottomrule
\end{tabular}
\caption{Token length statistics for claims and notes. Texts are tokenized by OpenAI's \href{https://github.com/openai/tiktoken}{tiktoken v0.11.0}.}
\label{tab:length_stats}
\end{table}

\subsection{Data Collection and Pre-processing}
We collected all our data from X's official \noteswebsite platform between January 2021 and December 2024, combining all publicly available releases. Because ratings are provided in multiple shards, we merge them and join them with the remaining components to obtain complete note–rating pairs. We then apply the official note-ranking algorithm to compute the aggregated helpfulness and reason labels. For helpfulness prediction, we remove entries labeled \texttt{NEED\_MORE\_RATINGS}, resulting in a binary label space--\texttt{CURRENTLY\_RATED\_HELPFUL} or \texttt{CURRENTLY\_RATED\_NOT\_HELPFUL}. The reason annotation is multi-label, covering 18 predefined categories. Finally, we retrieve tweet content and metadata via the X API and web crawling~\citep{borenstein-etal-2025-community}, linking each post with its corresponding notes to create the final dataset. See \Cref{subsec:preprocessing_details} for details of the preprocessing.

\Cref{tab:data_split} provides the data distribution. Our final dataset contains 104,966 posts, including 58,569 posts in English and 46,397 in other languages. Note that each post may be associated with one or more notes. We perform stratified partitioning with a ratio of 7:1:2 for training, development, and testing sets, maintaining separate splits for English and other language data. Henceforth we will refer to this dataset as \ourdataset.

\begin{table}[h]
\centering
\small
\begin{tabular}{ccc}
\toprule
\textbf{Notes per Post} & \textbf{Count (\%)} \\
\midrule
1 & 84.74\% \\
2 & 11.84\% \\
3 & \z2.43\% \\
4 & \z0.65\% \\
5 & \z0.18\% \\
6 & \z0.08\% \\
7 & \z0.03\% \\
8 & \z0.01\% \\
$\geq$9 & $<$0.02\% \\
\bottomrule
\end{tabular}
\caption{Distribution of the number of notes per post. The majority of posts have only one associated note.}
\label{tab:notes_per_claim}
\end{table}

\begin{figure*}[t]
    \centering
    \includegraphics[width=0.95\linewidth]{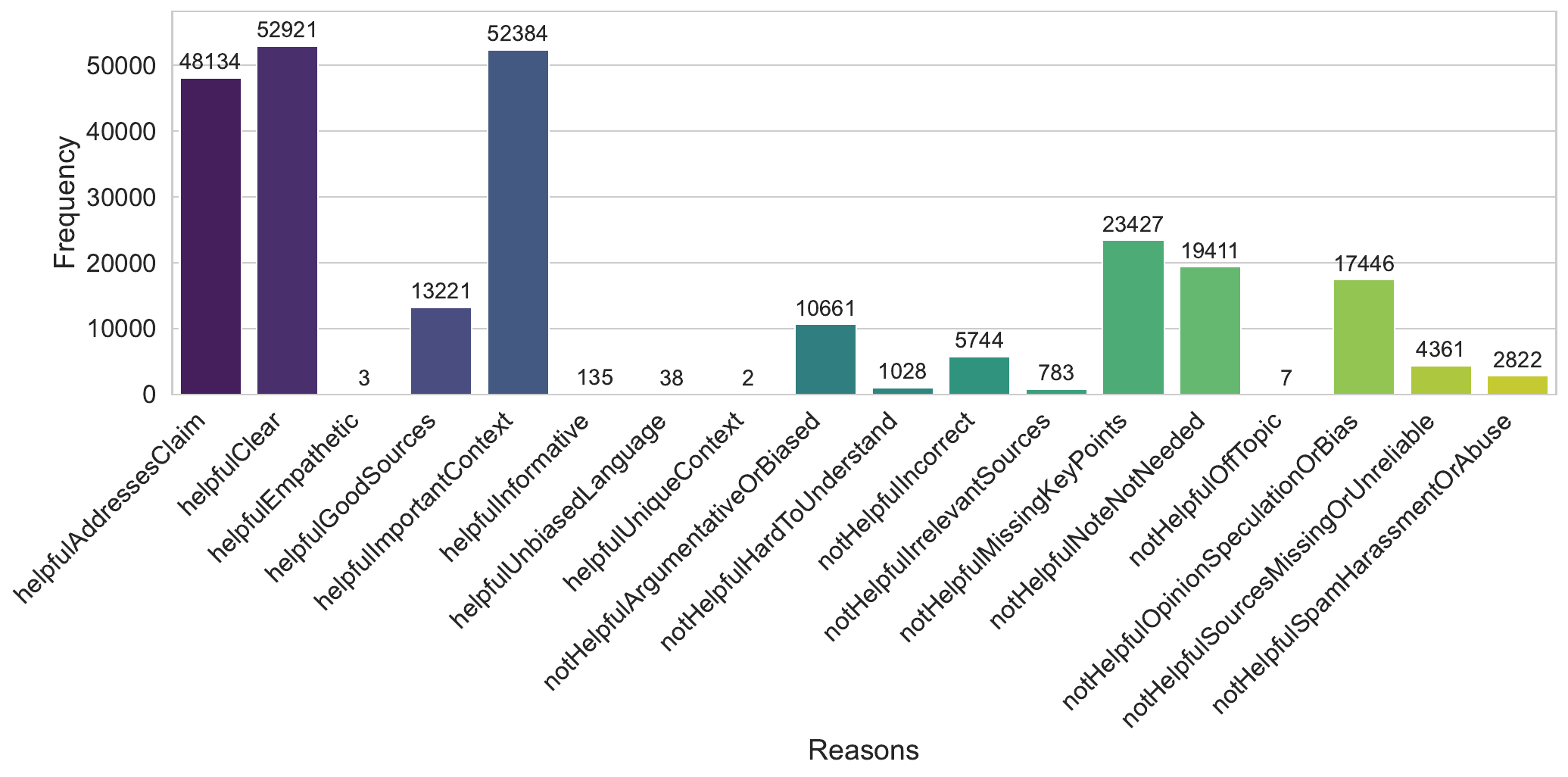}
    \caption{The histogram of the \emph{Note Reason} label distribution. \ourdataset contains 18 reason categories—8 corresponding to helpful notes and 10 to not helpful notes.}
    \label{fig:reason_dist}
\end{figure*}

\subsection{Dataset Statistics}
We inspected multiple aspects of our entire \ourdataset dataset. \Cref{tab:length_stats} shows the token length distribution for posts and notes correspondingly. The median token numbers are 42 for posts and 70 for notes, which suggests most posts and notes are short text snippets in our dataset. 

In \Cref{tab:notes_per_claim}, we observe that 84.74\% of posts have only one note, 11.84\% have two notes, and the remaining posts contain three or more notes. \Cref{tab:claim_helpfulness_distribution} shows that 64.36\% of posts have notes that are all rated as helpful, 31.30\% have only unhelpful notes, and the remaining 4.34\% include notes with mixed ratings. There are eight predefined reasons for a note being rated as helpful and ten reasons for being rated as unhelpful. The most common helpful reasons are \texttt{helpfulAddressClaim}, \texttt{helpfulImportantContext}, and \texttt{helpfulClear}, whereas \texttt{nothelpfulArgumentativeOrBiased} and \texttt{nothelpfulMissingKeyPoints} are the most frequent unhelpful reasons.

\begin{figure*}[t]
    \centering
    \includegraphics[width=\linewidth]{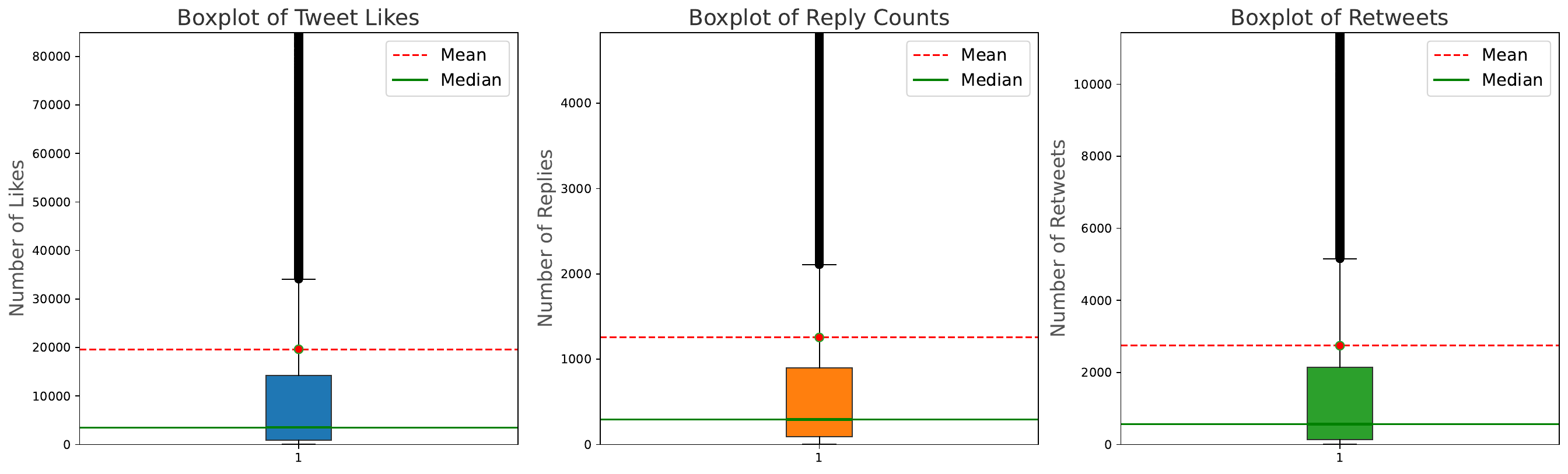}
    \caption{Boxplot of tweet post reactions in \ourdataset. We measure popularity using the number of replies, likes, and retweets. These posts receive a median of approximately 3k likes, 300 replies, and 700 retweets, indicating that notes are frequently attached to already popular content.}
    \label{fig:tweet_reactions}
\end{figure*}

\begin{figure}[t]
    \centering
    \includegraphics[width=0.9\linewidth]{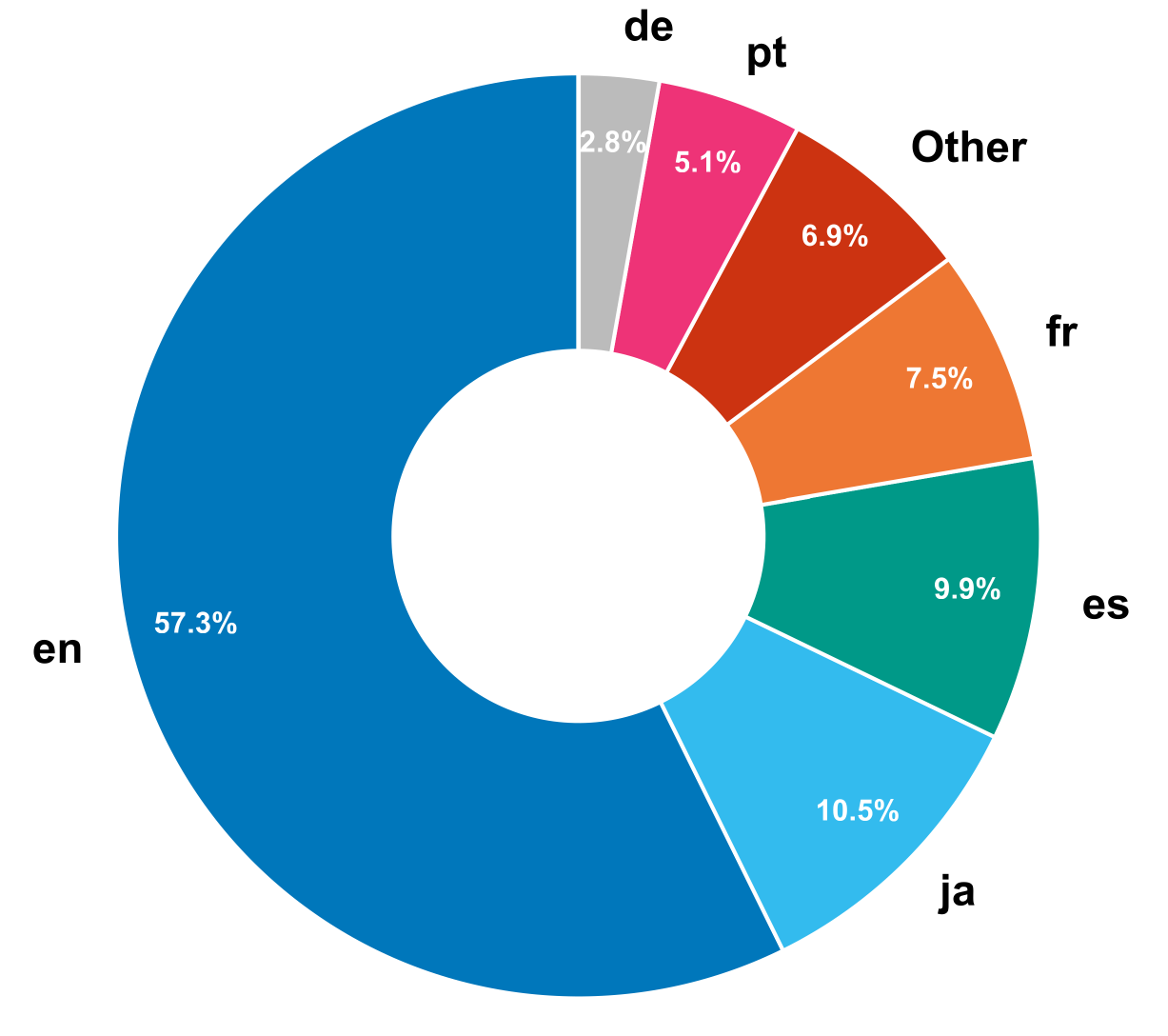}
    \caption{The language distribution of \ourdataset: en=English, fr=French, es=Spanish, ja=Japanese, pt=Portuguese, de=German, Other=languages that appear less than 1,000 times.}
    \label{fig:lang_dist}
\end{figure}

We present the distribution of helpfulness reasons in \ourdataset in \Cref{fig:reason_dist}. The most frequent helpful reason categories are \texttt{helpfulClear}, \texttt{helpfulImportantContext} and \texttt{helpfulAddressClaim}. They account for the majority of helpful instances in our dataset. 

\begin{table}[h]
\centering
\small
\begin{tabular}{lcc}
\toprule
\textbf{Post Type} & \textbf{Count} & \textbf{Percentage} \\
\midrule
All helpful notes & 67,558 & 64.36\% \\
All unhelpful notes & 32,851 & 31.30\% \\
Mixed helpful / unhelpful notes & \z4,557 & \z4.34\% \\
\bottomrule
\end{tabular}
\caption{Distribution of posts by helpfulness composition of their associated notes.}
\label{tab:claim_helpfulness_distribution}
\end{table}

The most frequent not helpful reasons in our dataset are \texttt{notHelpfulNoteNotNeeded}, \texttt{notHelpfulMissingKeyPoints}, and \texttt{notHelpfulOpinionSpeculationOrBias}. They form a distinct middle tier in \ourdataset. The remaining categories form a long tail with substantially lower frequencies.

Beyond note level statistics, we also examine the popularity of the posts targeted by notes in \Cref{fig:tweet_reactions}. We measure popularity using the number of replies, likes, and retweets. These posts receive a median of approximately 3k likes, 300 replies, and 700 retweets, indicating that notes are frequently attached to already popular content. As shown in \Cref{fig:tweet_reactions}, The mean reaction counts are substantial (e.g., $\approx$ 1,200 replies, $\approx$ 2,800 retweets, and $\approx$ 20,000 likes), indicating that posts in the dataset receive great public attention on average. However, the median for all three metrics is near zero. This large gap between the mean and median confirms the existence of a long tail of highly viral tweets. Given that such posts are flagged as potentially misleading, this highlights the importance of providing clear and helpful explanations at scale.

\Cref{fig:lang_dist} illustrates the language distribution of \ourdataset. English is the dominant language, accounting for 57.3\% of all notes, followed by Japanese (10.5\%) and Spanish (9.9\%). This indicates that our final dataset effectively captures the multilingual nature of crowd-sourced explanations.

\subsection{Benchmarking}
\paragraph{Experimental Setup} To establish a baseline for understanding the model performance in predicting the note helpfulness on \ourdataset, we conducted a comprehensive set of experiments using both Small Language Models (SLMs) (BERT-style encoder-only architectures in our paper) and decoder-only Large Language Models (LLMs). This setup provides a clear view of how different model architectures and capacities affect performance on our helpfulness evaluation task.

For SLMs, we formulate the task as multi-task classification, jointly predicting: (1) a binary helpfulness label; and (2) a multi-label reason classification, across 18 predefined categories. Inputs are constructed as a concatenated sequence in the format ``[CLS] Claim: \{claim\} [SEP] Note: \{note\_text\}''. For LLMs, we fine-tune using LoRA fine-tuning~\citep{hu2022lora} (see Appendix \Cref{tab:original_prompt} for detailed prompts).

We partitioned our experiments by model and language setting. For English, we experimented with SLMs including \bert~\citep{devlin-etal-2019-bert}, \modernbert~\citep{warner-etal-2025-smarter}, \roberta~\citep{liu2019robertarobustlyoptimizedbert} and \deberta (both base and large)~\citep{debertav3}, For multilingual evaluation, we used BERT-base-multilingual (\mbert) and \xlm. We also evaluated LLMs including \llama~\citep{llama3.1} and \mistral~\citep{mistral} for both English and multilingual settings on \ourdataset.

\begin{table*}[h!]
\centering
\small
\resizebox{2\columnwidth}{!}{
\begin{tabular}{lccccc}
\toprule
\textbf{Language} & \textbf{Model}  & \textbf{Helpfulness F1} & \textbf{Reason P} & \textbf{Reason R} & \textbf{Reason F1} \\
\midrule
\multirow{8}{3em}{\textit{English}} 
& \bertbase & 0.866 & 0.549 & 0.742 & 0.631 \\
& \bertlarge & 0.874 & 0.553 & 0.758 & 0.640 \\
& \robertabase & 0.876 & 0.546 & 0.783 & 0.643 \\
& \robertalarge & 0.890 & 0.581 & 0.772 & 0.663 \\
& ModernBERT-base & 0.871 & 0.545 & 0.755 & 0.633 \\
& ModernBERT-large & 0.888 & 0.576 & 0.752 & 0.652 \\
& \debertabase & 0.886 & 0.541 & \textbf{0.810} & 0.650 \\
& \debertalarge & 0.896 & 0.589 & 0.759 & \textbf{0.665} \\
& \llama & 0.919 & \textbf{0.640} & 0.640 & 0.640 \\
& \mistral & \textbf{0.920} & 0.620 & 0.620 & 0.620  \\
\midrule
\multirow{2}{3em}{\textit{Other}} 
& \mbert& 0.873 & 0.562 & 0.799 & 0.659 \\
& XLM-RoBERTa-base & 0.882 & 0.584 & 0.798 & 0.674 \\
& XLM-RoBERTa-large & 0.894 & 0.603 & \textbf{0.815} & \textbf{0.693} \\
& \llama & \textbf{0.928} & \textbf{0.653} & 0.652 & 0.653 \\
& \mistral & 0.926 & 0.647 & 0.647 & 0.647  \\
\bottomrule
\end{tabular}
}
\caption{Benchmark results for helpfulness and reasons prediction. \emph{Helpfulness} means the binary classification of notes helpfulness. \emph{Reason} means multi-label classification of helpfulness reasons. \emph{P} is Precision and \emph{R} is Recall.}
\label{tab:benchmark_results}
\end{table*}

\paragraph{Results and Analysis} As shown in \Cref{tab:benchmark_results}, models perform well on helpfulness prediction, with most of them achieving F1 above 0.88. This suggests that this task is relatively easy. In contrast, reason prediction is much more challenging, as reflected by substantially lower F1 for all models, which consistently fall below 0.70.

Comparing SLMs to LLMs: \Cref{tab:benchmark_results} shows that LLMs such as \llama and \mistral outperform smaller SLMs on helpfulness prediction. In paticular, \mistral achieves the highest F1 score (0.920), demonstrating the advantages of larger models. However, LLMs perform worse than SLMs in reason prediction, with F1 remaining comparable to those of the best-performing SLMs such as \debertalarge. 

Within the SLM families (BERT, RoBERTa, DeBERTa), larger models consistently outperform their base-sized counterparts. For example, \robertalarge achieves better scores than \robertabase across all metrics. This highlights the importance of model capacity for capturing the nuances required for this complex task. The table also shows that \deberta models generally perform at or near the top among the SLMs, particularly for the reasons task, with the highest F1 Score of 0.665 among all SLMs.

The multilingual part show a similar trend: larger models achieve strong performance in Helpfulness prediction while smaller models like mBert and XLM-RoBERTa exhibit better performance on the reason prediction. Among all models, XLM-RoBERTa-large achieves the highest Reason F1 of 0.693. This indicates that these models are effective at generalizing across different languages.

\begin{figure*}[tb]
\centering
\includegraphics[width=1\textwidth,scale=1.5]{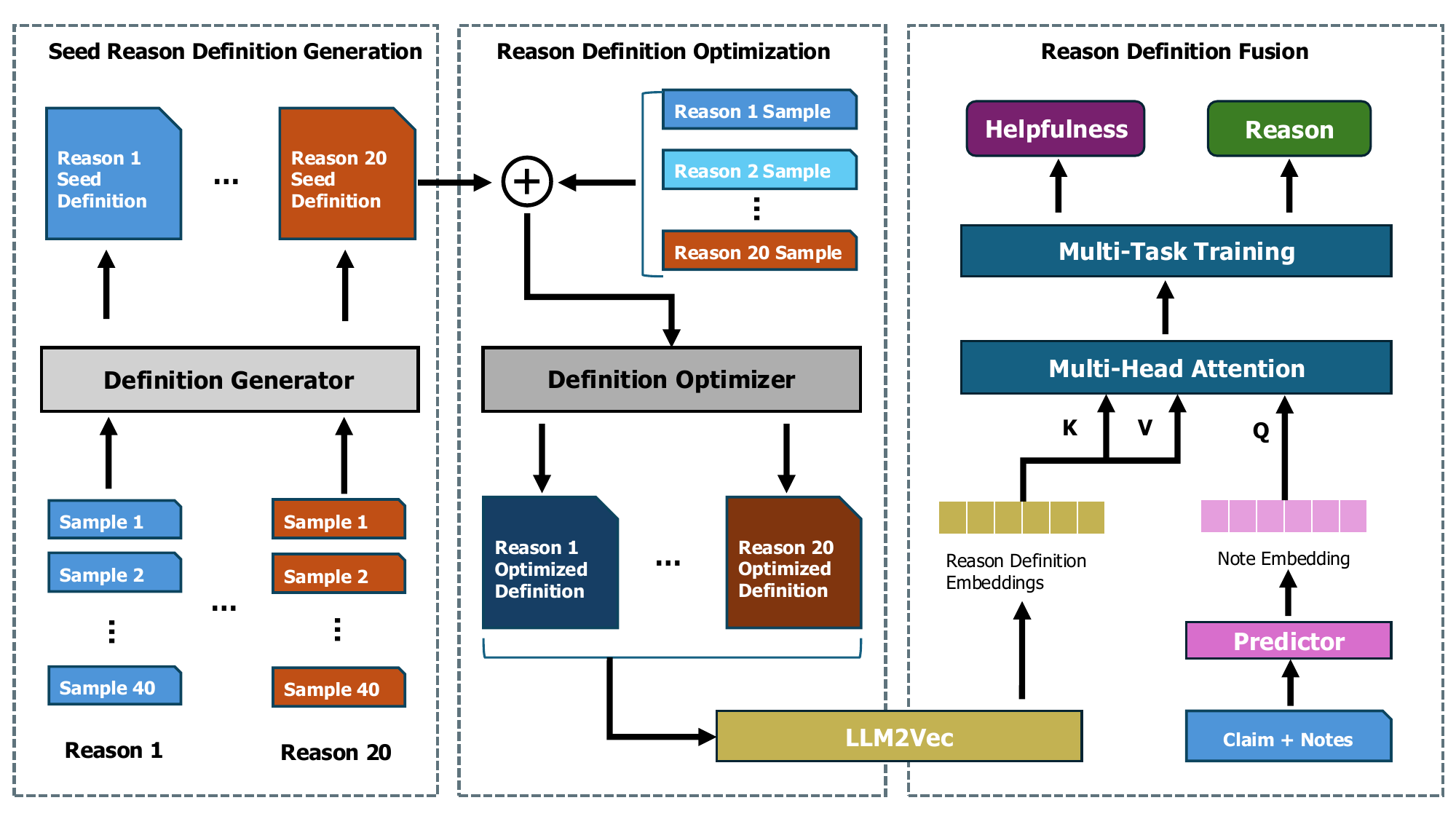}
\caption{Our framework for automatic reason definition generation and optimization. (1) Seed reason definitions are generated from sampled post-note pairs. (2) The definitions are refined via \promptagent. (3) The optimized definitions are embedded and fused with note representations for joint helpfulness and reason prediction.}
\label{fig:framework}
\end{figure*}

\section{Enhancing Helpfulness Prediction with Reason Definitions}
Preliminary analysis shows that predicting why a particular explanation is (un)helpful is often subjective across instances, largely due to the lack of unified or official definitions. This subjectivity poses a barrier to model performance and limits the explainability of community-based fact-checking platforms. Prior work has demonstrated that providing clear label definitions can significantly improves the classification performance~\citep{gao-etal-2023-benefits,peskine-etal-2023-definitions,khatuya-etal-2025-label}. 

Given the large number of notes with annotated reasons available, we aim to automatically generate and optimize reason label definitions based on data. To this end, we introduce a pipeline that leverages Automatic Prompt Optimization (APO) to generate and improve helpfulness reason definitions, which are then integrated with notes for reason prediction. Since the majority of notes are written in English, we focus on \textbf{English} data for the remainder of our experiments. \Cref{fig:framework} illustrates the overall pipeline of our approach.

\subsection{Automatically Optimizing Helpfulness Reason Definition}
We automatically generate and optimize reason definitions following the first two steps in \Cref{fig:framework}: (1) \textit{Seed reason definition generation}; and (2) \textit{Reason definition optimization}. In the first step, we follow~\citet{peskine-etal-2023-definitions} and randomly sample 40 instances per helpfulness reason category. We then prompt the \gptfouro model\footnote{Detailed model version: \href{https://platform.openai.com/docs/models/gpt-4o}{gpt-4o-2024-08-06}.} to generate an initial ``seed'' version of the definitions based on the sampled candidate instances (see Appendix \Cref{tab:gen_def} for detailed prompts to generate seed definitions).\footnote{Since a single sample  may have multiple reason labels, we perform an additional check to ensure that each sample appears only within the same category to encourage diversity.} In the second step, we employ \promptagent~\citep{wang2024promptagent}, a Monte Carlo Tree Search (MCTS)-based automatic prompt optimization framework, to refine the initial definitions. 

\promptagent adopts an iterative refinement process that emulates how human experts strategically craft effective prompts, using an MCTS-based planning mechanism. Each refined prompt corresponds to a new state in the search space, obtained by the refiner agent model through feedback collected from the feedback agent model. \promptagent relies on the MCTS state–action value function to estimate the reward of each state. MCTS iteratively performs four key operations—selection, expansion, simulation, and backpropagation—until a predefined number of iterations is reached. The highest-reward trajectory is then selected to produce the final refined prompt. We initialize the process with the seed definitions obtained from step~(1) in~\Cref{fig:framework} and run adapted \promptagent using a definition-refinement feedback loop.

\begin{table}[h!]
\centering
\resizebox{1\columnwidth}{!}{
\begin{tabular}{lcccc}
\toprule
\textbf{Model} & \textbf{Helpfulness F1} & \textbf{Reason F1} \\
\midrule
Llama3.1-8B-ins & 0.636 & 0.126 \\
Llama3.1-8B-ins-seed  & 0.783 & 0.304 \\
Llama3.1-8B-ins-opt & \textbf{0.785} & \textbf{0.344} \\
\midrule % Optional rule to separate model groups
Mistral-7B-ins  & 0.761 & 0.135  \\
Mistral-7B-ins-seed & 0.776 & 0.257 \\
Mistral-7B-ins-opt & \textbf{0.786} & \textbf{0.324}  \\
\bottomrule
\end{tabular}
}
\caption{Zero-shot prediction performance for LLMs w/o adding reason definitions on our \ourdataset~test set. \emph{No suffix} means the vanilla model prediction performance. \emph{seed} means definitions are generated by \gptfouro on randomly sampled cases. \emph{opt} means that the definitions are further optimized by \promptagent.}
\label{tab:add_def_results}
\end{table}

To evaluate the impact of incorporating generated definitions, we perform zero-shot prompting with \llama and \mistral for both helpfulness and reason prediction. \Cref{tab:add_def_results} reports results for the vanilla setting, as well as for models integrated with seed and optimized definitions. We observe substantial F1 improvements in both helpfulness and reason prediction when adding seed definitions (on average around 15\%), with further gains after definition optimization, indicating that refined definitions provide additional benefits.

For \llama, both seed and optimized definitions yield large improvements. Although \mistral starts stronger than \llama in the vanilla setting, it still benefits notably from incorporating definitions, particularly in reason F1. These results suggest that incorporating reason definition information substantially enhances model performance—not only in predicting whether a given explanation note is helpful, but also in identifying the underlying reason with respect to the claim. (See detailed seed definition in \Cref{tab:seed_def} and optimized reason definition in \Cref{tab:optimized_def} in the Appendix).

\begin{table}[h!]
\centering
\resizebox{1\columnwidth}{!}{
\begin{tabular}{lccc}
\toprule
\textbf{Model} & \textbf{Helpfulness F1} & \textbf{Reason F1} \\
\midrule
\robertabase & 0.876 & 0.643 \\
\textbf{\robertabase-MHA} & \textbf{0.887} & \textbf{0.652} \\
\midrule
\robertalarge & 0.890 & 0.663 \\
\textbf{\robertalarge-MHA} & \textbf{0.899} & \textbf{0.672} \\
\midrule
\modernbertbase & 0.871 & 0.633 \\
\textbf{\modernbertbase-MHA} & \textbf{0.874} & \textbf{0.644} \\
\midrule
\modernbertlarge & 0.888 & 0.652 \\
\textbf{\modernbertlarge-MHA} & \textbf{0.898} & \textbf{0.662} \\
\midrule
\debertabase & 0.886 & 0.650 \\
\textbf{\debertabase-MHA} & \textbf{0.887} & \textbf{0.668} \\
\midrule
\debertalarge & 0.896 & 0.665 \\
\textbf{\debertalarge-MHA} & 0.896 & \textbf{0.677} \\
\bottomrule
\end{tabular}
}
\caption{Reason prediction performance comparison of base SLMs versus fusing optimized reason definition with the Multi-Head Attention module.}
\label{tab:mha_results}
\end{table}

\subsection{Training with Reason Definition Fusion}
After obtaining the optimized helpfulness reason definitions, we proceed to the third step in \Cref{fig:framework}, where we incorporate this information into the training SLM predictors for helpfulness and reason prediction. The core idea is that the predictor should attend more strongly to the most relevant helpfulness reasons for a given claim–note pair.

We obtain note embeddings $E_{note}$ from the base predictors and reason embeddings $E_{reason}$ from the optimized reason definitions using \textsc{LLM2Vec}~\citep{llm2vec}.\footnote{Specifically, we use the following model: \href{https://huggingface.co/McGill-NLP/LLM2Vec-Meta-Llama-31-8B-Instruct-mntp-supervised}{\texttt{LLM2Vec-Meta-Llama-31-8B-Instruct-mntp-supervised}}.} We then apply a Multi-Head Attention (MHA) mechanism, using $E_{note}$ as the query (Q) and $E_{reason}$ as the key (K) and value (V), to fuse the reason information into the note representation. Finally, we adopt a multi-task learning objective that jointly optimizes helpfulness prediction and multi-label reason classification for our experiments.

\Cref{tab:mha_results} shows that integrating optimized reason definitions via MHA consistently improves performance across all SLMs. For example, \robertalarge-MHA achieves a Helpfulness F1 of 0.899 and a Reason F1 of 0.669, outperforming its vanilla counterpart (0.890 and 0.663, respectively). Similar trends hold for \modernbert and \deberta, with \debertalarge-MHA achieving the highest Reason F1 score of 0.677. 

Overall, these results demonstrate that attention-based fusion of reason definitions effectively enhances SLMs for both helpfulness and reason prediction on \ourdataset.

\begin{table}[h!]
\centering
\resizebox{1\columnwidth}{!}{
\begin{tabular}{lccccc}
\toprule
\textbf{Predictors} & \textbf{NEI P} & \textbf{NEI R} & \textbf{NEI F1} \\
\midrule
\robertabase-FT & 0.869 & 0.920 & 0.894 \\
\robertabase-notes & 0.742 & 0.521 & 0.612 \\
\robertalarge-FT & 0.882 & \textbf{0.935} & \textbf{0.908} \\
\robertalarge-notes & 0.731 & 0.676 & 0.702 \\
\modernbertbase-FT & 0.845 & 0.843 & 0.844 \\
\modernbertbase-notes & 0.770 & 0.779 & 0.774 \\
\modernbertlarge-FT & 0.902 & 0.906 & 0.904 \\
\modernbertlarge-notes & 0.808 & 0.807 & 0.808 \\
\debertabase-FT & 0.912 & 0.855 & 0.882 \\
\debertabase-notes & 0.746 & 0.553 & 0.635 \\
\debertalarge-FT & \textbf{0.919} & 0.890 & 0.905 \\
\debertalarge-notes & 0.736 & 0.774 & 0.755 \\
\bottomrule
\end{tabular}
}
\caption{Predictor generalization performance on evidence sufficiency task. \textit{notes} stands for models for original communitynotes predictors. \textit{FT} means \textbf{F}ine\textbf{T}uned models on evidence sufficiency task. \textbf{NEI} means \textbf{N}ot \textbf{E}nough \textbf{I}nformation label.}
\label{tab:evi_suf_results}
\end{table}

\section{Generalization of the Explanation Helpfulness}
In this section, we assess the generalization and the utility of our helpfulness predictors beyond community notes. A key challenge is that most existing datasets lack explicit annotations for explanation of helpfulness. Therefore, we explore two key scenarios in automated fact-checking. First, we evaluate whether helpfulness prediction can generalize to evidence sufficiency prediction, which is an important subtask in fact-checking~\citep{atanasova-etal-2022-fact}. Second, we explore the effect of incorporating evidence helpfulness in automated fact-checking~\citep{xing-etal-2025-evaluating}. 

\subsection{Generalization to Evidence Sufficiency}
\label{subsec:evi_suf}
Evidence sufficiency is a task that decides whether the information in the evidence is sufficient for predicting the veracity of a claim. Automated fact-checking models can make veracity predictions only when there is enough evidence~\citep{atanasova-etal-2022-fact}. In this scenario, we would like to explore whether our helpfulness predictors can predict the sufficiency of an evidence (since a helpful evidence must be ``sufficient''). We adopted \emph{Sufficient Fact}~\citep{atanasova-etal-2022-fact}, a diagnostic fact-checking dataset to detect when evidence with omitted information is (in)sufficient. To adapt our prediction, we treat sufficient evidence (\textit{EI-ENOUGH INFO}) as helpful and insufficient evidence (\textit{NEI-NOT ENOUGH INFO}) as unhelpful. \Cref{tab:evi_suf_results} shows the model performance on the evidence sufficiency task, focusing on the \textit{NEI} label. 

As expected, fine-tuned (FT) models outperform the original notes predictors, with \robertalarge-FT achieving the best \textit{NEI} F1 of 0.908. Nevertheless, the notes models show competitive performance: for example, \modernbertlarge-notes achieves a \textit{NEI} F1 of 0.808, only about 10\% lower than its FT counterpart. Across all predictors, the top predicted reason is \texttt{notHelpfulMissingKeyPoints}, which closely corresponds to the \textit{NEI} label, as both signify the absence of critical information. This indicates that helpfulness prediction carries transferable signal for evidence sufficiency. 

Still, helpfulness and sufficiency are not the same: even sufficient evidence could be predicted as unhelpful for other reasons such as \texttt{notHelpfulArgumentativeOrBiased}, which may account for the observed performance gap in our experiments.

\subsection{Incorporating Evidence Helpfulness in Automated Fact Checking}
In this part, we further investigate whether predicted evidence helpfulness can enhance automated fact-checking. Helpfulness scores provide auxiliary signals for assessing evidence quality, with the potential to help models prioritize informative evidence during prediction. We adopted \climatefever~\citep{diggelmann2020climatefever}, a real-world dataset for verifying climate change–related claims. Each claim is associated with multiple evidence sentences, to which we apply our predictors to assign helpfulness scores and reasons. To perform fact-checking, we use \llama to perform fact-checking (see detailed prompt in Appendix \Cref{tab:fc_prompt}) prompting the model with both the claim and its associated evidence. 

We compare the F1 with and without the evidence helpfulness information and report results in \Cref{tab:fc_results}. Overall, incorporating helpfulness improves fact-checking performance, with \modernbertbase achieving 0.537 accuracy and \modernbert and \robertabase improving performance by around 2\%. While \modernbert demonstrates major gains, \deberta and \roberta yield smaller, non-significant improvements. Interestingly, this pattern mirrors their performance drop in cross-domain evidence sufficiency evaluation in \Cref{subsec:evi_suf} (despite strong in-domain results in \Cref{tab:benchmark_results}). One possible explanation is that models overfit \ourdataset and generalize less effectively to other domains.

\begin{table}[h!]
\centering
\small
\resizebox{1\columnwidth}{!}{
\begin{tabular}{lcc}
\toprule
\textbf{Predictors} & \textbf{F1} & \textbf{+Helpfulness F1} \\
\midrule
\robertabase & 0.517 & \textbf{0.521} \\
\robertalarge & 0.519 & 0.521 \\
\modernbertbase & 0.521 & \textbf{0.535} \\
\modernbertlarge & 0.517 & \textbf{0.535} \\
\debertabase & 0.517 & 0.519 \\
\debertalarge & 0.518 & 0.519 \\
\bottomrule
\end{tabular}
}
\caption{Automated fact-checking performance of \llama on the \climatefever dataset. \emph{F1} stands for Accuracy. \emph{+Helpfulness F1} means accuracy incorporating helpfulness information. Numbers in bold represents the gap between \emph{+Helpfulness} and basic prediction accuracy is statistically significant (using a two-tailed t-test with $p<0.05$).}
\label{tab:fc_results}
\end{table}

\section{Conclusion}
In this paper, we introduced the novel task of predicting the helpfulness of explanations and their underlying reasons for a given claim. To support this task, we constructed \ourdataset, a large-scale post–note helpfulness dataset based on X's \noteswebsite, and established benchmarks with pretrained language models. 

While existing models perform reasonably well on overall helpfulness, they struggle to identify underlying reasons. To address this limitation, we proposed a framework that automatically generates and optimizes reason definitions, which, when reintegrated, greatly improves reason prediction. We further demonstrated that helpfulness prediction can help predict evidence sufficiency and enhance automated fact-checking systems.

Beyond performance gains, our findings highlight the importance of making evaluation criteria for explanation helpfulness more explicit and structured. We hope that our dataset and framework can serve as a foundation for building stronger helpfulness predictors capable of assigning prior labels to real-time community notes, thereby streamlining the community-based fact-checking process. Additionally, the automatically generated and optimized reason definitions may enhance the explainability and trustworthiness of automated fact-checking models and could be extended to improve other classification tasks beyond this domain. 

In future work, we plan to extend our framework to multilingual and multimodal settings and evaluate it in more advanced end-to-end fact-checking systems and real-world applications.

\section*{Limitations}
While our study provides new insights into the helpfulness of crowd-sourced fact-checking explanations, several limitations remain.

First, the inherently subjective nature of community-based annotations introduces potential noise and inconsistency in the helpfulness and reason labels. Despite our efforts to aggregate ratings through official mechanisms, differences in annotator interpretation and cultural context may still influence the results.

Second, our automated definition generation and optimization pipeline has thus far been evaluated primarily on English. Although our dataset includes multilingual content, we did not systematically validate the quality or transferability of the generated definitions across languages. Future work should explore multilingual adaptation and cross-lingual alignment of reason definitions.

Third, while our experiments show improved model performance when integrating optimized reason definitions, the evaluation is limited to classification metrics. Further human-centered evaluation is necessary to assess whether these improvements translate to greater interpretability and practical usefulness in real-world fact-checking workflows.

Finally, our study focuses on publicly available \emph{Community Notes} data, which may reflect platform-specific norms and biases. For a broader generalization to other fact-checking ecosystems (e.g., Meta's or TikTok's community moderation systems), we need further investigation.

\section*{Ethics and Broader Impact}
Our work aims to improve the transparency, explainability, and scalability of community-based fact-checking systems. By modeling and predicting the helpfulness of user-generated explanations, we hope to assist both contributors and platforms in identifying notes that are more informative, balanced, and useful for the public discourse.

\paragraph{Bias and Fairness.}
Community-based ratings may reflect social, cultural, or political biases that influence which notes are deemed “helpful.” Our models inevitably inherit such biases from the underlying data. We would like to emphasize that model predictions should not be directly used to automatically moderate or suppress user content. Instead, they should be interpreted as auxiliary signals to guide human review.

\paragraph{Responsible Use.}
While our framework can help prioritize potentially helpful notes, it is not intended to replace human judgment. Automated systems that assign helpfulness or credibility labels at scale may affect content visibility, user reputation, and online discourse. We encourage incorporating fairness-aware and interpretable mechanisms before deployment.

\paragraph{Positive Societal Impact.}
Despite these caveats, our dataset and models can help combat misinformation by improving the quality, trustworthiness, and accessibility of community fact-checking. Making helpfulness definitions explicit and machine-interpretable supports more transparent and accountable information verification.

\paragraph{Data Collection and Licenses}
All data that we used in this work originated from the publicly available \textit{X Community Notes} dataset,\footnote{\url{https://communitynotes.x.com/guide/en/under-the-hood/download-data}}
 which was released under the X platform's open data policy. The dataset contains user-generated notes, associated posts, and helpfulness and reason labels, without any personally identifiable information. We release only anonymized text and aggregated statistics in compliance with the X platform's terms of service. The dataset is multilingual, with English, Japanese, and Spanish being the best represented languages. Any future release of derived data will follow ACL data ethics guidelines with appropriate documentation, to enable transparent, reproducible research in NLP and computational social science.

\paragraph{Security Implication}
Our study introduces no direct security risks, but models trained on user-generated content may be vulnerable to misuse, such as large-scale manipulation or suppression of dissent. To mitigate these risks, we release only aggregated and anonymized data and emphasize that our models are intended for research use, not automated moderation or deployment without human oversight. Future work should address adversarial attacks, bias exploitation, and model inversion risks when integrating such models into live platforms.

\section*{Acknowledgments}
We thank our reviewers for their valuable reviews and feedback, which greatly contributed to the improvement of our paper.

\bibliography{custom}

\begin{thebibliography}{35}
\providecommand{\natexlab}[1]{#1}

\bibitem[{Atanasova et~al.(2020)Atanasova, Simonsen, Lioma, and Augenstein}]{atanasova-etal-2020-generating-fact}
Pepa Atanasova, Jakob~Grue Simonsen, Christina Lioma, and Isabelle Augenstein. 2020.
\newblock \href {https://doi.org/10.18653/v1/2020.acl-main.656} {Generating fact checking explanations}.
\newblock In \emph{Proceedings of the 58th Annual Meeting of the Association for Computational Linguistics}, ACL~2020, pages 7352--7364, Online. Association for Computational Linguistics.

\bibitem[{Atanasova et~al.(2022)Atanasova, Simonsen, Lioma, and Augenstein}]{atanasova-etal-2022-fact}
Pepa Atanasova, Jakob~Grue Simonsen, Christina Lioma, and Isabelle Augenstein. 2022.
\newblock \href {https://doi.org/10.1162/tacl_a_00486} {Fact checking with insufficient evidence}.
\newblock \emph{Transactions of the Association for Computational Linguistics}, 10:746--763.

\bibitem[{Augenstein et~al.(2025)Augenstein, Bakker, Chakraborty, Corney, Ferrara, Gurevych, Hale, Hovy, Ji, Larraz, Menczer, Nakov, Papotti, Sahnan, Warren, and Zagni}]{communitymoderationnewepistemology}
Isabelle Augenstein, Michiel Bakker, Tanmoy Chakraborty, David Corney, Emilio Ferrara, Iryna Gurevych, Scott Hale, Eduard Hovy, Heng Ji, Irene Larraz, Filippo Menczer, Preslav Nakov, Paolo Papotti, Dhruv Sahnan, Greta Warren, and Giovanni Zagni. 2025.
\newblock \href {https://arxiv.org/abs/2505.20067} {Community moderation and the new epistemology of fact checking on social media}.
\newblock \emph{ArXiv preprint}, abs/2505.20067.

\bibitem[{BehnamGhader et~al.(2024)BehnamGhader, Adlakha, Mosbach, Bahdanau, Chapados, and Reddy}]{llm2vec}
Parishad BehnamGhader, Vaibhav Adlakha, Marius Mosbach, Dzmitry Bahdanau, Nicolas Chapados, and Siva Reddy. 2024.
\newblock \href {https://openreview.net/forum?id=IW1PR7vEBf} {{LLM2V}ec: Large language models are secretly powerful text encoders}.
\newblock In \emph{Proceedings of the First Conference on Language Modeling}, COLM~24, Philadelphia, PA, USA.

\bibitem[{Borenstein et~al.(2025)Borenstein, Warren, Elliott, and Augenstein}]{borenstein-etal-2025-community}
Nadav Borenstein, Greta Warren, Desmond Elliott, and Isabelle Augenstein. 2025.
\newblock \href {https://doi.org/10.18653/v1/2025.acl-short.42} {Can community notes replace professional fact-checkers?}
\newblock In \emph{Proceedings of the 63rd Annual Meeting of the Association for Computational Linguistics}, ACL~2025, pages 535--552, Vienna, Austria. Association for Computational Linguistics.

\bibitem[{Chuai et~al.(2024)Chuai, Pilarski, Renault, Restrepo-Amariles, Troussel-Clément, Lenzini, and Pröllochs}]{chuai_community_2024}
Yuwei Chuai, Moritz Pilarski, Thomas Renault, David Restrepo-Amariles, Aurore Troussel-Clément, Gabriele Lenzini, and Nicolas Pröllochs. 2024.
\newblock \href {https://arxiv.org/abs/2409.08781} {Community-based fact-checking reduces the spread of misleading posts on social media}.
\newblock \emph{ArXiv preprint}, abs/2409.08781.

\bibitem[{De et~al.(2025)De, Bakker, Baxter, and Saveski}]{supernotes}
Soham De, Michiel~A. Bakker, Jay Baxter, and Martin Saveski. 2025.
\newblock \href {https://doi.org/10.1145/3696410.3714934} {Supernotes: Driving consensus in crowd-sourced fact-checking}.
\newblock In \emph{Proceedings of the ACM on Web Conference 2025}, WWW~25, page 3751–3761, Sydney NSW, Australia. Association for Computing Machinery.

\bibitem[{Devlin et~al.(2019)Devlin, Chang, Lee, and Toutanova}]{devlin-etal-2019-bert}
Jacob Devlin, Ming-Wei Chang, Kenton Lee, and Kristina Toutanova. 2019.
\newblock \href {https://doi.org/10.18653/v1/N19-1423} {{BERT}: Pre-training of deep bidirectional transformers for language understanding}.
\newblock In \emph{Proceedings of the 2019 Conference of the North {A}merican Chapter of the Association for Computational Linguistics: Human Language Technologies}, NAACL~2019, pages 4171--4186, Minneapolis, Minnesota. Association for Computational Linguistics.

\bibitem[{Diggelmann et~al.(2020)Diggelmann, Boyd-Graber, Bulian, Ciaramita, and Leippold}]{diggelmann2020climatefever}
Thomas Diggelmann, Jordan Boyd-Graber, Jannis Bulian, Massimiliano Ciaramita, and Markus Leippold. 2020.
\newblock \href {https://arxiv.org/abs/2012.00614} {{CLIMATE-FEVER}: A dataset for verification of real-world climate claims}.
\newblock \emph{ArXiv preprint}, abs/2012.00614.

\bibitem[{Drolsbach and Pr\"{o}llochs(2023)}]{diffusionnotes}
Chiara~Patricia Drolsbach and Nicolas Pr\"{o}llochs. 2023.
\newblock \href {https://doi.org/10.1145/3610058} {Diffusion of community fact-checked misinformation on {Twitter}}.
\newblock In \emph{Proceedings of the ACM on Human-Computer Interaction}, New York, USA. Association for Computing Machinery.

\bibitem[{Dubey et~al.(2024)Dubey, Jauhri, Pandey, Kadian, Al{-}Dahle, Letman, Mathur, Schelten, Yang, Fan, Goyal, Hartshorn, Yang, Mitra, Sravankumar, Korenev, Hinsvark, Rao, Zhang, Rodriguez, Gregerson, Spataru, Rozi{\`{e}}re, Biron, Tang, Chern, and et~al.}]{llama3.1}
Abhimanyu Dubey, Abhinav Jauhri, Abhinav Pandey, Abhishek Kadian, Ahmad Al{-}Dahle, Aiesha Letman, Akhil Mathur, Alan Schelten, Amy Yang, Angela Fan, Anirudh Goyal, Anthony Hartshorn, Aobo Yang, Archi Mitra, Archie Sravankumar, Artem Korenev, Arthur Hinsvark, Arun Rao, Aston Zhang, and 8 others. 2024.
\newblock \href {https://arxiv.org/abs/2407.21783} {The {L}lama 3 herd of models}.
\newblock \emph{ArXiv preprint}, abs/2407.21783.

\bibitem[{Eldifrawi et~al.(2024)Eldifrawi, Wang, and Trabelsi}]{eldifrawi-etal-2024-automated}
Islam Eldifrawi, Shengrui Wang, and Amine Trabelsi. 2024.
\newblock \href {https://doi.org/10.18653/v1/2024.acl-long.361} {Automated justification production for claim veracity in fact checking: A survey on architectures and approaches}.
\newblock In \emph{Proceedings of the 62nd Annual Meeting of the Association for Computational Linguistics}, ACL~2024, pages 6679--6692, Bangkok, Thailand. Association for Computational Linguistics.

\bibitem[{Fernando et~al.(2024)Fernando, Banarse, Michalewski, Osindero, and Rockt{\"{a}}schel}]{promptbreeder}
Chrisantha Fernando, Dylan Banarse, Henryk Michalewski, Simon Osindero, and Tim Rockt{\"{a}}schel. 2024.
\newblock \href {https://openreview.net/forum?id=9ZxnPZGmPU} {Prompt{B}reeder: Self-referential self-improvement via prompt evolution}.
\newblock In \emph{Proceedings of Forty-first International Conference on Machine Learning}, ICML~2024, Vienna, Austria. OpenReview.net.

\bibitem[{Gao et~al.(2023)Gao, Ghosh, and Gimpel}]{gao-etal-2023-benefits}
Lingyu Gao, Debanjan Ghosh, and Kevin Gimpel. 2023.
\newblock \href {https://doi.org/10.18653/v1/2023.emnlp-main.853} {The benefits of label-description training for zero-shot text classification}.
\newblock In \emph{Proceedings of the 2023 Conference on Empirical Methods in Natural Language Processing}, EMNLP~2023, pages 13823--13844, Singapore. Association for Computational Linguistics.

\bibitem[{Guo et~al.(2022)Guo, Schlichtkrull, and Vlachos}]{guo-etal-2022-survey}
Zhijiang Guo, Michael Schlichtkrull, and Andreas Vlachos. 2022.
\newblock \href {https://doi.org/10.1162/tacl_a_00454} {A survey on automated fact-checking}.
\newblock \emph{Transactions of the Association for Computational Linguistics}, 10:178--206.

\bibitem[{He et~al.(2023)He, Gao, and Chen}]{debertav3}
Pengcheng He, Jianfeng Gao, and Weizhu Chen. 2023.
\newblock \href {https://openreview.net/pdf?id=sE7-XhLxHA} {{DeBERTaV3}: Improving {DeBERTa} using {ELECTRA}-style pre-training with gradient-disentangled embedding sharing}.
\newblock In \emph{Proceedings of The Eleventh International Conference on Learning Representations}, ICLR~2023, Kigali, Rwanda. OpenReview.net.

\bibitem[{Hu et~al.(2022)Hu, Shen, Wallis, Allen{-}Zhu, Li, Wang, Wang, and Chen}]{hu2022lora}
Edward~J. Hu, Yelong Shen, Phillip Wallis, Zeyuan Allen{-}Zhu, Yuanzhi Li, Shean Wang, Lu~Wang, and Weizhu Chen. 2022.
\newblock \href {https://openreview.net/forum?id=nZeVKeeFYf9} {Lo{RA}: Low-rank adaptation of large language models}.
\newblock In \emph{Proceedings of the Tenth International Conference on Learning Representations}, ICLR~2022, Virtual Event. OpenReview.net.

\bibitem[{Jiang et~al.(2023)Jiang, Sablayrolles, Mensch, Bamford, Chaplot, de~Las~Casas, Bressand, Lengyel, Lample, Saulnier, Lavaud, Lachaux, Stock, Scao, Lavril, Wang, Lacroix, and Sayed}]{mistral}
Albert~Q. Jiang, Alexandre Sablayrolles, Arthur Mensch, Chris Bamford, Devendra~Singh Chaplot, Diego de~Las~Casas, Florian Bressand, Gianna Lengyel, Guillaume Lample, Lucile Saulnier, L{\'{e}}lio~Renard Lavaud, Marie{-}Anne Lachaux, Pierre Stock, Teven~Le Scao, Thibaut Lavril, Thomas Wang, Timoth{\'{e}}e Lacroix, and William~El Sayed. 2023.
\newblock \href {https://arxiv.org/abs/2310.06825} {Mistral 7b}.
\newblock \emph{ArXiv preprint}, abs/2310.06825.

\bibitem[{Khatuya et~al.(2025)Khatuya, Naidu, Ghosh, Goyal, and Ganguly}]{khatuya-etal-2025-label}
Subhendu Khatuya, Shashwat Naidu, Saptarshi Ghosh, Pawan Goyal, and Niloy Ganguly. 2025.
\newblock \href {https://doi.org/10.18653/v1/2025.findings-acl.1145} {Label-semantics aware generative approach for domain-agnostic multilabel classification}.
\newblock In \emph{Findings of the Association for Computational Linguistics: ACL 2025}, ACL~2025, pages 22286--22298, Vienna, Austria. Association for Computational Linguistics.

\bibitem[{Kumar et~al.(2025)Kumar, Venkata, Khandelwal, Santra, Agrawal, and Gupta}]{kumar-etal-2025-sculpt}
Shanu Kumar, Akhila~Yesantarao Venkata, Shubhanshu Khandelwal, Bishal Santra, Parag Agrawal, and Manish Gupta. 2025.
\newblock \href {https://doi.org/10.18653/v1/2025.acl-long.730} {{SCULPT}: Systematic tuning of long prompts}.
\newblock In \emph{Proceedings of the 63rd Annual Meeting of the Association for Computational Linguistics}, ACL~2025, pages 14996--15029, Vienna, Austria. Association for Computational Linguistics.

\bibitem[{Lewandowsky et~al.(2012)Lewandowsky, Ecker, Seifert, Schwarz, and Cook}]{stephan-misinformation-2012}
Stephan Lewandowsky, Ullrich K.~H. Ecker, Colleen~M. Seifert, Norbert Schwarz, and John Cook. 2012.
\newblock \href {https://doi.org/10.1177/1529100612451018} {Misinformation and its correction: Continued influence and successful debiasing}.
\newblock \emph{Psychological Science in the Public Interest}, 13(3):106--131.
\newblock PMID: 26173286.

\bibitem[{Liu et~al.(2019)Liu, Ott, Goyal, Du, Joshi, Chen, Levy, Lewis, Zettlemoyer, and Stoyanov}]{liu2019robertarobustlyoptimizedbert}
Yinhan Liu, Myle Ott, Naman Goyal, Jingfei Du, Mandar Joshi, Danqi Chen, Omer Levy, Mike Lewis, Luke Zettlemoyer, and Veselin Stoyanov. 2019.
\newblock \href {https://arxiv.org/abs/1907.11692} {{RoBERTa}: A robustly optimized {BERT} pretraining approach}.
\newblock \emph{ArXiv preprint}, abs/1907.11692.

\bibitem[{Martel and Rand(2024)}]{Martel2024FactcheckerWL}
Cameron Martel and David~G. Rand. 2024.
\newblock \href {https://api.semanticscholar.org/CorpusID:272334488} {Fact-checker warning labels are effective even for those who distrust fact-checkers.}
\newblock \emph{Nature Human Behaviour}, 8:1957 -- 1967.

\bibitem[{Nakov et~al.(2021)Nakov, Corney, Hasanain, Alam, Elsayed, Barr{\'{o}}n{-}Cede{\~{n}}o, Papotti, Shaar, and Martino}]{afc_nakov_2021}
Preslav Nakov, David P.~A. Corney, Maram Hasanain, Firoj Alam, Tamer Elsayed, Alberto Barr{\'{o}}n{-}Cede{\~{n}}o, Paolo Papotti, Shaden Shaar, and Giovanni Da~San Martino. 2021.
\newblock \href {https://doi.org/10.24963/IJCAI.2021/619} {Automated fact-checking for assisting human fact-checkers}.
\newblock In \emph{Proceedings of the Thirtieth International Joint Conference on Artificial Intelligence}, IJCAI~2021, pages 4551--4558, Montreal, Canada.

\bibitem[{Peskine et~al.(2023)Peskine, Koren{\v{c}}i{\'c}, Grubisic, Papotti, Troncy, and Rosso}]{peskine-etal-2023-definitions}
Youri Peskine, Damir Koren{\v{c}}i{\'c}, Ivan Grubisic, Paolo Papotti, Raphael Troncy, and Paolo Rosso. 2023.
\newblock \href {https://doi.org/10.18653/v1/2023.findings-emnlp.267} {Definitions matter: Guiding {GPT} for multi-label classification}.
\newblock In \emph{Findings of the Association for Computational Linguistics}, EMNLP~2023, pages 4054--4063, Singapore. Association for Computational Linguistics.

\bibitem[{Ramnath et~al.(2025)Ramnath, Zhou, Guan, Mishra, Qi, Shen, Wang, Woo, Jeoung, Wang, Wang, Ding, Lu, Xu, Zhou, Srinivasan, Yan, Chen, Ding, Xu, and Cheong}]{ramnath2025systematicsurveyautomaticprompt}
Kiran Ramnath, Kang Zhou, Sheng Guan, Soumya~Smruti Mishra, Xuan Qi, Zhengyuan Shen, Shuai Wang, Sangmin Woo, Sullam Jeoung, Yawei Wang, Haozhu Wang, Han Ding, Yuzhe Lu, Zhichao Xu, Yun Zhou, Balasubramaniam Srinivasan, Qiaojing Yan, Yueyan Chen, Haibo Ding, and 2 others. 2025.
\newblock \href {https://arxiv.org/abs/2502.16923} {A systematic survey of automatic prompt optimization techniques}.
\newblock \emph{ArXiv preprint}, abs/2502.16923.

\bibitem[{Renault et~al.(2024)Renault, Restrepo-Amariles, and Troussel}]{renault_collaboratively_2024}
Thomas Renault, David Restrepo-Amariles, and Aurore Troussel. 2024.
\newblock \href {https://doi.org/10.2139/ssrn.4800565} {Collaboratively adding context to social media posts reduces the sharing of false news}.
\newblock HEC Research Papers Series 1519, HEC Paris.

\bibitem[{Russo et~al.(2023)Russo, Tekiro{\u{g}}lu, and Guerini}]{russo-etal-2023-benchmarking}
Daniel Russo, Serra~Sinem Tekiro{\u{g}}lu, and Marco Guerini. 2023.
\newblock \href {https://doi.org/10.1162/tacl_a_00601} {Benchmarking the generation of fact checking explanations}.
\newblock \emph{Transactions of the Association for Computational Linguistics}, 11:1250--1264.

\bibitem[{Wang et~al.(2024)Wang, Li, Wang, Bai, Luo, Zhang, Jojic, Xing, and Hu}]{wang2024promptagent}
Xinyuan Wang, Chenxi Li, Zhen Wang, Fan Bai, Haotian Luo, Jiayou Zhang, Nebojsa Jojic, Eric~P. Xing, and Zhiting Hu. 2024.
\newblock \href {https://openreview.net/forum?id=22pyNMuIoa} {Prompt{A}gent: Strategic planning with language models enables expert-level prompt optimization}.
\newblock In \emph{Proceedings of the International Conference on Learning Representations}, ICLR~2024, Vienna, Austria. OpenReview.net.

\bibitem[{Warner et~al.(2025)Warner, Chaffin, Clavi{\'e}, Weller, Hallstr{\"o}m, Taghadouini, Gallagher, Biswas, Ladhak, Aarsen, Adams, Howard, and Poli}]{warner-etal-2025-smarter}
Benjamin Warner, Antoine Chaffin, Benjamin Clavi{\'e}, Orion Weller, Oskar Hallstr{\"o}m, Said Taghadouini, Alexis Gallagher, Raja Biswas, Faisal Ladhak, Tom Aarsen, Griffin~Thomas Adams, Jeremy Howard, and Iacopo Poli. 2025.
\newblock \href {https://doi.org/10.18653/v1/2025.acl-long.127} {Smarter, better, faster, longer: A modern bidirectional encoder for fast, memory efficient, and long context finetuning and inference}.
\newblock In \emph{Proceedings of the 63rd Annual Meeting of the Association for Computational Linguistics}, ACL~2025, pages 2526--2547, Vienna, Austria. Association for Computational Linguistics.

\bibitem[{Warren et~al.(2025)Warren, Shklovski, and Augenstein}]{explainablefc}
Greta Warren, Irina Shklovski, and Isabelle Augenstein. 2025.
\newblock \href {https://doi.org/10.1145/3706598.3713277} {Show me the work: Fact-checkers' requirements for explainable automated fact-checking}.
\newblock In \emph{Proceedings of the 2025 CHI Conference on Human Factors in Computing Systems}, CHI~2025, Yokohama, Japan. Association for Computing Machinery.

\bibitem[{Wojcik et~al.(2022)Wojcik, Hilgard, Cortex, Judd, Mocanu, Ragain, Hunzaker, Product, and Baxter}]{Wojcik2022BirdwatchCW}
Stefan Wojcik, Sophie Hilgard, Twitter Cortex, Nick Judd, Delia Mocanu, Stephen Ragain, M.B.~Fallin Hunzaker, Keith Coleman~Twitter Product, and Jay Baxter. 2022.
\newblock \href {https://arxiv.org/abs/2210.15723} {Birdwatch: Crowd wisdom and bridging algorithms can inform understanding and reduce the spread of misinformation}.
\newblock \emph{ArXiv preprint}, abs/2210.15723.

\bibitem[{Xie et~al.(2025)Xie, Xing, Wang, Geng, Iqbal, Sahnan, Gurevych, and Nakov}]{xie-etal-2025-fire}
Zhuohan Xie, Rui Xing, Yuxia Wang, Jiahui Geng, Hasan Iqbal, Dhruv Sahnan, Iryna Gurevych, and Preslav Nakov. 2025.
\newblock \href {https://doi.org/10.18653/v1/2025.findings-naacl.158} {{FIRE}: Fact-checking with iterative retrieval and verification}.
\newblock In \emph{Findings of the Association for Computational Linguistics}, NAACL~2025, pages 2901--2914, Albuquerque, New Mexico. Association for Computational Linguistics.

\bibitem[{Xing et~al.(2025)Xing, Baldwin, and Lau}]{xing-etal-2025-evaluating}
Rui Xing, Timothy Baldwin, and Jey~Han Lau. 2025.
\newblock \href {https://doi.org/10.18653/v1/2025.naacl-long.282} {Evaluating evidence attribution in generated fact checking explanations}.
\newblock In \emph{Proceedings of the 2025 Conference of the Nations of the Americas Chapter of the Association for Computational Linguistics: Human Language Technologies}, NAACL~2025, pages 5475--5496, Albuquerque, New Mexico. Association for Computational Linguistics.

\bibitem[{Xing et~al.(2022)Xing, Bhatia, Baldwin, and Lau}]{xing-etal-2022-automatic}
Rui Xing, Shraey Bhatia, Timothy Baldwin, and Jey~Han Lau. 2022.
\newblock \href {https://aclanthology.org/2022.alta-1.16/} {Automatic explanation generation for climate science claims}.
\newblock In \emph{Proceedings of the 20th Annual Workshop of the Australasian Language Technology Association}, ALTA~2022, pages 122--129, Adelaide, Australia.

\end{thebibliography}

\newpage
\appendix

\section{Appendix}
\label{sec:appendix}

\subsection{Extended Background of X Community Notes Program}
\paragraph{Origin} X~\noteswebsite originated from the Birdwatch~\citep{Wojcik2022BirdwatchCW} program which was launched by X (formerly Twitter) in 2021.\footnote{\href{https://blog.x.com/en_us/topics/product/2021/introducing-birdwatch-a-community-based-approach-to-misinformation}{Introducing Birdwatch, a community-based approach to misinformation, By Keith Coleman, Monday, 25 January 2021.}} Participants, known as contributors, write “notes” that clarify or contextualize claims made in posts, and other users rate these notes for helpfulness and accuracy. Notes receiving broad agreement across diverse raters are surfaced publicly beneath the corresponding posts. This community-driven system aims to promote transparency and provide multiple perspectives in addressing misinformation on social media~\citep{communitymoderationnewepistemology}. Initially limited to U.S. users, the program gained wider attention in 2022 during major misinformation events such as the Russian invasion of Ukraine and the COVID-19 pandemic. Birdwatch was re-branded as \noteswebsite and expanded globally in November 2022. As of November 2023, the program had over 133,000 contributors, with notes reportedly viewed tens of millions of times per day, reflecting its growing role in mitigating misinformation in online social media.

\paragraph{Advantages and Challenges} \noteswebsite introduces a promising new paradigm for fact-checking. It democratizes content moderation by enabling the public to collectively determine whether a note should be added to a post, offering a less intrusive alternative to traditional expert-based fact-checking systems that often rely on explicit warning labels. This community-driven approach also accelerates the detection of misinformation, offering better scalability. However, \noteswebsite also faces several challenges. The community can still favor popularity over truth, similar to the echo-chamber effects often seen on social media. While its scalability and faster operation are promising, studies show that its performance still lags behind expert-based fact-checking in many cases~\citep{Martel2024FactcheckerWL}. In addition, helpful notes often appear too late—after a post has already received significant engagement—reducing their effectiveness in correcting misinformation and changing user beliefs~\citep{communitymoderationnewepistemology, supernotes, borenstein-etal-2025-community}.

\begin{table*}[ht]
\small
\centering
\begin{tabularx}{\textwidth}{@{}llX@{}}
\toprule
\textbf{Phase} & \textbf{\#} & \textbf{Step Description} \\
\midrule
Prescoring & 1. & Pre-filter Data: Include only raters ($\ge 10$ ratings) \& notes ($\ge 5$ ratings). Coalesce similar raters. \\
           & 2. & Initial Model Fit: Fit matrix factorization (MF) models for each scorer. Assign intermediate note status based on intercept thresholds. \\
           & 3. & Calculate \& Filter Helpfulness: Compute Author/Rater Helpfulness Scores from the initial MF results. Filter out low-helpfulness raters. \\
           & 4. & Refine Score: Fit a tag-consensus MF model on the helpfulness-filtered data and update all helpfulness scores. \\
\midrule
Scoring    & 1. & Load \& Refresh Data: Load prescoring outputs. Re-run the pre-filtering step on the newest available data. \\
           & 2. & Re-fit Main Models: Re-fit the main MF models (for all scorers) using the helpfulness-filtered ratings. \\
           & 3. & Fit Diligence Model: Fit the note diligence MF model. \\
           & 4. & Compute Confidence Bounds: Add pseudo-ratings and re-fit models to get upper/lower confidence bounds for note intercepts. \\
           & 5. & Reconcile \& Finalize Status: Generate a final status for each note from all scorer results. Stabilize status for notes older than two weeks. \\
           & 6. & Assign Explanation Tags: Assign the top two matching explanation tags. If two such tags are not found, revert status to \texttt{NEED\_MORE\_RATINGS}. \\
\bottomrule
\end{tabularx}
\caption{Overview of the X \noteswebsite note ranking Algorithm pipeline.}
\label{tab:ranking_algo}
\end{table*}

\subsection{\ourdataset Pre-processing Details}\label{subsec:preprocessing_details}
\paragraph{Raw Data} Community notes data are publicly available on the official website.\footnote{\url{https://x.com/i/communitynotes/download-data}} The data is released in several following separate files:
\begin{itemize}
    \item \textbf{Notes:} This table contains all notes and their attributes. Core attributes include \textit{tweetId}, which identifies the tweet the note is associated with, and \textit{classification}, indicating whether the tweet is considered misleading. It also includes several attributes describing potential types of misinformation (e.g., \textit{misleadingFactualError}); however, these fields are not directly related to note ratings and are excluded from this study.
    \item \textbf{Ratings:} This table includes all user ratings of notes. Our helpfulness label scheme is derived from this table, where attributes prefixed with \textit{helpful-} and \textit{notHelpful-} correspond to the helpfulness reasons used in our \ourdataset experiments.
    \item \textbf{Note Status History:} This table contains metadata about notes, including their status updates and timestamps. We use the \textit{currentStatus} attribute to determine each note’s final helpfulness label.
    \item \textbf{User Enrollment:} This table provides metadata on each user’s enrollment status in the Community Notes program. Since it does not contain information related to note helpfulness, we exclude it from our experiments.
    \item \textbf{Note Requests:} This table records all requests submitted by X users for a Community Note. It was added after our data collection period, so it is not used in this study.
\end{itemize}

We first construct the full dataset by joining all data files on the \textit{noteID} column. We then remove empty notes and merge the duplicated helpfulness reasons \texttt{notHelpfulOpinionSpeculation} and \texttt{notHelpfulOpinionSpeculationOrBias} into a single reason category which is \texttt{notHelpfulOpinionSpeculationOrBias}. Finally, we exclude notes that are labeled as \texttt{notHelpfulOther}, as their reasons are unclear and uninformative. 

\begin{table}[h]
\centering
\small
\resizebox{1\columnwidth}{!}{
\begin{tabular}{lcccccc}
\toprule
\textbf{Language} & \multicolumn{3}{c}{\textbf{Before Processing}} & \multicolumn{3}{c}{\textbf{After Processing}} \\
\cmidrule(lr){2-4} \cmidrule(lr){5-7}
 & \textbf{Train} & \textbf{Dev} & \textbf{Test} & \textbf{Train} & \textbf{Dev} & \textbf{Test} \\
\midrule
English & 41,102 & 5,872 & 11,744 & 40,994 & 5,858 & 11,717 \\
Other & 32,520 & 4,646 & 9,292 & 32,478 & 4,638 & 9,281 \\
\midrule
\textbf{Total} & 73,622 & 10,518 & 21,036 & 73,472 & 10,496 & 20,998 \\
\bottomrule
\end{tabular}
}
\caption{Comparison of dataset sizes before and after processing for English and other language subsets.}
\label{tab:before_after_processing}
\end{table}

\paragraph{Note Ranking Algorithm}
\Cref{tab:ranking_algo} summarizes the ranking algorithm used by \noteswebsite. Contributors submit notes for potentially misleading posts, which are rated by other users and aggregated to assign one of three statuses --- \texttt{CURRENTLY\_RATED\_HELPFUL}, \texttt{CURRENTLY\_RATED\_NOT\_HELPFUL}, or \texttt{NEED\_MORE\_RATINGS} --- determining public visibility. All notes initially receive the \texttt{NEED\_MORE\_RATINGS} and remain unpublished until they accumulate at least five ratings.

The ranking algorithm is based on a Matrix Factorization (MF) model trained on a sparse note--rater matrix, producing a continuous \emph{Note Helpfulness Score} for each note. A note is assigned \texttt{CURRENTLY\_RATED\_HELPFUL} if its score exceeds 0.40; currently, only notes associated with potentially misleading posts and meeting this threshold are shown publicly. Conversely, a note is assigned \texttt{CURRENTLY\_RATED\_NOT\_HELPFUL} if its score falls below $-0.05 - 0.8 \times |\text{noteFactorScore}|$, or if its upper confidence bound is smaller than $-0.04$. Notes with scores between these thresholds remain at \texttt{NEED\_MORE\_RATINGS} state until additional ratings are collected. \Cref{tab:before_after_processing} presents data size statistics for English and other language subsets before and after processing in \ourdataset.

\begin{table}[ht]
\centering
\small
\resizebox{1\columnwidth}{!}{
\begin{tabular}{lrr}
\toprule
\textbf{Reason Category} & \textbf{Support} & \textbf{F1} \\ 
\midrule
\textit{Top 3 Predicted Reasons} & & \\
helpfulAddressesClaim & 6,630 & 0.794 \\
helpfulImportantContext & 5,222 & 0.713 \\
notHelpfulArgumentativeOrBiased & 1,454 & 0.658 \\
\midrule
\textit{Bottom 3 Predicted Reasons} & & \\
notHelpfulIncorrect & 780 & 0.345 \\
notHelpfulHardToUnderstand & 156 & 0.306 \\
notHelpfulIrrelevantSources & 84 & 0.096 \\
\bottomrule
\end{tabular}
}
\caption{\debertalarge F1 broken down.}
\label{tab:reason_f1_breakdown}
\end{table}

\subsection{Further Analysis of SLM performance }
As reason prediction is much more challenging compared with helpfulness prediction, in order to better understand the task, we further analyze model F1 scores across different reason categories. We chose \deberta, which achieves the best reason prediction performance in the benchmark results (\Cref{tab:benchmark_results}). We report the top and bottom five reason categories by F1 in \Cref{tab:reason_f1_breakdown}. The model achieves its highest F1 scores on frequent categories such as \texttt{helpfulAddressesClaim} (0.794) and \texttt{helpfulImportantContext} (0.713). Conversely, it struggles with sparse categories such as \texttt{helpfulEmpathetic} and \texttt{helpfulInformative} (both 0.00 F1), highlighting a long-tail challenge where the model fails to generalize to infrequent reasons with limited training examples. Notably, there is systematic confusion among \texttt{helpfulClear}, \texttt{helpfulImportantContext}, and \texttt{helpfulAddressesClaim}. This suggests that while the model can reliably identify whether a note is helpful, it often struggles to distinguish fine-grained semantic differences between closely related reason categories.

\begin{table}[h]
\resizebox{1\columnwidth}{!}{
\centering
\small
\begin{tabular}{lcc}
\toprule
\textbf{Model} & \textbf{Training Epoch} & \textbf{Batch Size} \\
\midrule
\bertbase & 19 & 64 \\
\bertlarge & 8 & 32 \\
\robertabase & 16 & 64 \\
\robertalarge & 10 & 32 \\
\modernbertbase & 6 & 32 \\
\modernbertlarge & 5 & 16 \\
\debertabase & 10 & 32 \\
\debertalarge & 8 & 16 \\
\xlmbase & 20 & 64 \\
\xlmlarge & 10 & 32 \\
\mbert & 11 & 64 \\
\bottomrule
\end{tabular}
}
\caption{Training Epoch and Batch Size for all SLMs used in our experiments on \ourdataset.}
\label{tab:slm_config}
\end{table}

\subsection{Experimental Details}
We report the best training epochs and batch sizes for SLMs in \Cref{tab:slm_config}. To reduce overfitting, we use smaller batch sizes and fewer epochs for base models, and larger ones for their large counterparts. For LLMs, we apply LoRA fine-tuning with a rank of 16, alpha of 32, and applied to all linear modules. We train with a batch size of 4, gradient accumulation of 1 step, for 3 epochs, using a cosine learning rate scheduler with a warmup ratio of 0.1. The training takes between 32 and 46 GPU hours on 2 NVIDIA RTX6000 Ada Generation GPUs (48GB GPU RAM). All open source models are directly adopted from the Hugging Face Hub, including SLMs (\bert\footnote{\url{https://huggingface.co/docs/transformers/en/model_doc/bert}}, \roberta\footnote{\url{https://huggingface.co/docs/transformers/en/model_doc/roberta}}, \deberta\footnote{\url{https://huggingface.co/docs/transformers/en/model_doc/deberta}}, \modernbert\footnote{\url{https://huggingface.co/docs/transformers/en/model_doc/modernbert}}, \mbert\footnote{\url{https://huggingface.co/google-bert/bert-base-multilingual-cased}}, \xlm\footnote{\url{https://huggingface.co/docs/transformers/en/model_doc/xlm-roberta}}) and LLMs (\llama\footnote{\url{https://huggingface.co/meta-llama/Llama-3.1-8B-Instruct}}, \mistral\footnote{\url{https://huggingface.co/mistralai/Mistral-7B-Instruct-v0.3}}).

We reimplemented the complete ranking algorithm on a 64-core Intel(R) Xeon(R) Gold 6448H CPU with 500~GB of RAM to recover aggregated helpfulness labels and reason annotations for each note in \noteswebsite.\footnote{\url{https://github.com/twitter/communitynotes}} 

\subsection{Prompts}
In this section, we provided detailed prompts used in our \ourdataset experiments. \Cref{tab:gen_def} shows the prompt for initial reason generation. \Cref{tab:original_prompt} presents the prompt for LLM to predict Helpfulness and Reasons. \Cref{tab:seed_def_prompt} shows English prompts used for helpfulness and reason prediction on \ourdataset with seed definition. \Cref{tab:optimized_prompt} shows the same task with optimized definition for LLM predictions. \Cref{tab:seed_def} provides the seed definition generated using \gptfouro over random samples. \Cref{tab:optimized_def} is the detailed reason definition optimized using \promptagent. \Cref{tab:fc_prompt} shows the prompt we used for LLM to perform fact checking on \climatefever dataset.

\clearpage
\onecolumn
\footnotesize

% prompt for reason definition generation
\begin{longtable}{p{1\columnwidth}}
\toprule
\textbf{Initial reason generation} \\
\midrule
\endfirsthead

\toprule
\textbf{Initial reason generation} \\
\midrule
\endhead

\midrule 
\multicolumn{1}{r}{{Continued on next page}} \\
\endfoot

\bottomrule
\\[\abovecaptionskip]
\caption{Prompt for initial reason generation.}
\label{tab:gen_def}
\endlastfoot
You will be given a set of samples, each sample contains CLAIM, their corresponding NOTE to explain the CLAIM. All samples provided are \$\{helpful\_label\} in explaining the CLAIM and associated with the same REASON. Your task is to conclude the definition of the REASON.

Here are samples:
\$\{samples\}

The REASON for above samples being \$\{helpful\_label\} is \$\{reason\_label\}. After checking these samples, the definition of this REASON is:\\
\end{longtable}

% original prompt for helpfulness prediction
\begin{longtable}{p{1\columnwidth}}
\toprule
\textbf{Original Prompt} \\
\midrule
\endfirsthead

\toprule
\textbf{Original Prompt} \\
\midrule
\endhead

\midrule 
\multicolumn{1}{r}{{Continued on next page}} \\
\endfoot

\bottomrule
\\[\abovecaptionskip]
\caption{Original English prompt for note helpfulness and reason prediction.}
\label{tab:original_prompt}
\endlastfoot
Given a potentially misleading CLAIM and an associated NOTE, your task is to determine whether the NOTE is helpful in clarifying the CLAIM and identify two reasons from the predefined reason set explaining why it is helpful or not helpful. \newline

The predefined reason set is:{'helpfulAddressesClaim', 'helpfulClear', 'helpfulEmpathetic', 'helpfulGoodSources', 'helpfulImportantContext', 'helpfulInformative', 'helpfulUnbiasedLanguage', 'helpfulUniqueContext', 'notHelpfulArgumentativeOrBiased', 'notHelpfulHardToUnderstand','notHelpfulIncorrect', 'notHelpfulIrrelevantSources', 'notHelpfulMissingKeyPoints', 'notHelpfulNoteNotNeeded', 'notHelpfulOffTopic', 'notHelpfulOpinionSpeculationOrBias', 'notHelpfulSourcesMissingOrUnreliable', 'notHelpfulSpamHarassmentOrAbuse'}\newline

Output in the following JSON format only, no extra output:
\{"helpfulness": helpful or non\_helpful,"reasons":"reason1;reason2"\} 
CLAIM: \$\{claim\}\newline
NOTE: \$\{note\}\newline
Answer: \\
\end{longtable}

% seed definition prompt for reason prediction
\begin{longtable}{p{1\columnwidth}}
\toprule
\textbf{Prompt with seed defininion} \\
\midrule
\endfirsthead

\toprule
\textbf{Prompt with seed definition} \\
\midrule
\endhead

\midrule 
\multicolumn{1}{r}{{Continued on next page}} \\
\endfoot

\bottomrule
\\[\abovecaptionskip]
\caption{English prompt with seed definition for note helpfulness and reason prediction.}
\label{tab:seed_def_prompt}
\endlastfoot
Given a potentially misleading CLAIM and an associated NOTE, your task is to determine whether the NOTE is helpful in clarifying the CLAIM and identify two reasons from the predefined reason set explaining why it is helpful or not helpful. \newline

Here are reasons and their definitions:
\$\{reason definitions\}

Output in the following JSON format only, no extra output:
\{"helpfulness": helpful or non\_helpful,"reasons":"reason1;reason2"\}\newline
CLAIM: \$\{claim\}\newline
NOTE: \$\{note\}\newline              
Answer: \\
\end{longtable}

% optimized prompt for reason prediction
\begin{longtable}{p{1\columnwidth}}
\toprule
\textbf{Optimized Prompt} \\
\midrule
\endfirsthead

\toprule
\textbf{Optimized Prompt} \\
\midrule
\endhead

\midrule 
\multicolumn{1}{r}{{Continued on next page}} \\
\endfoot

\bottomrule
\\[\abovecaptionskip]
\caption{Optimized English prompt for note helpfulness and reason prediction.}
\label{tab:optimized_prompt}
\endlastfoot

For each claim-note pair below, select exactly two reasons (from the provided list) that most accurately explain whether and why the note is helpful or not helpful for understanding or resolving the claim. \\
\textbf{High-Level Criteria}\newline
\textbf{1. Prioritizing Central, Accurate Clarification}
\begin{itemize}[leftmargin=*,topsep=2pt,noitemsep]
    \item Assign a “helpful” reason only if the note is factually accurate, well-supported, and directly clarifies, corrects, or provides essential context for the main assertion, a key factual sub-claim, or the foundational evidence of the claim. Peripheral or partially related details are insufficient.
    \item If the note’s chief function is to correct a specific fact, figure, identity, date, source, or a significant misattribution or misconception central to the claim, this counts as clarification or correction (use “helpfulClear” and/or “helpfulAddressesClaim”).
    \item Never assign any “helpful” reason if the note contains substantive factual errors, misrepresentations, speculation, or bias regarding any key claim point or related evidence—regardless of any attempt to clarify.
\end{itemize} 
\textbf{2. “Helpful” Reason Prioritization}
\begin{itemize}[leftmargin=*,topsep=2pt,noitemsep]
    \item helpfulClear and helpfulAddressesClaim take precedence over other “helpful” reasons when a note offers a direct factual correction or explicit clarification for the claim’s core point or supporting evidence.
    \item helpfulGoodSources should only be selected in addition if the note’s correction is fundamentally grounded in clearly cited, authoritative, and faithfully summarized sources (not just the presence of a source).
    \item helpfulImportantContext is used when the note supplies background or context without which the claim would be misunderstood or misinterpreted, and this context is not a direct factual correction but provides essential interpretive clarity.
    \item Use helpfulUnbiasedLanguage or helpfulEmpathetic only in conjunction with a substantive clarification or correction and if the neutrality or tone of the explanation is materially helpful. 
    When both direct correction and crucial context are present, prefer (b) helpfulClear or (a) helpfulAddressesClaim as primary, paired with (e) helpfulImportantContext as secondary only if context is indispensable and not redundant with the correction.
    \item helpfulGoodSources never substitutes for correction—pick it only if the faithful use of sources is a main reason for helpfulness. 
\end{itemize} \\
\textbf{3. “NotHelpful” Reason Selection} 
\begin{itemize}[leftmargin=*,topsep=2pt,noitemsep]
    \item Assign “notHelpfulIncorrect” if there is any inaccuracy, factual misstatement, or misleading claim regarding the main point or evidence.
    \item Assign “notHelpfulMissingKeyPoints” if the note avoids or omits addressing the core issue or supporting fact, no matter how detailed its peripheral info.
    \item Use “notHelpfulNoteNotNeeded” if the note is trivial, redundant, or supplies information already evident and non-essential for claim comprehension.
    \item “notHelpfulSourcesMissingOrUnreliable” or “notHelpfulIrrelevantSources” apply if sources are not credible, are misrepresented, or are irrelevant to the core of the claim.
    \item If tone, personal opinion, bias, speculation, or argumentativeness blocks any clarification, use one of the corresponding “notHelpful” reasons that best fits the limitation.
    \item For unclear, incomplete, or confusing notes, use “notHelpfulHardToUnderstand.”
    “notHelpfulOther” and “notHelpfulOffTopic” only if none of the above describe the problem or the note is wholly irrelevant. 
\end{itemize} 
\textbf{Reason Definitions}:\$\{Reason Definitions\} \\
\textbf{Decision Process Checklist} \newline
1. Is the note factually accurate and not misleading about any major point or evidence?
\begin{itemize}[leftmargin=*,topsep=2pt,noitemsep]
    \item If no, assign “notHelpfulIncorrect” and another as fits.
    \item If yes, proceed.
\end{itemize}
2. Does the note directly clarify, correct, or critically contextualize the main assertion, a key factual sub-claim, or any central supporting evidence?
\begin{itemize}[leftmargin=*,topsep=2pt,noitemsep]
    \item If yes, select the most specific “helpful” reason(s) per above priority order.
    \item If its primary value is sources, include “helpfulGoodSources” only if the sourcing itself is decisive.
    \item Do not select “helpfulImportantContext” unless the info is both necessary for accurate interpretation and not primarily a direct correction.
    \item If note only offers peripheral detail, trivia, or sidesteps the key issue, use “notHelpfulMissingKeyPoints” and/or “notHelpfulNoteNotNeeded.”
\end{itemize}
3. Is the note clear, neutral, and respectful in tone?
\begin{itemize}[leftmargin=*,topsep=2pt,noitemsep]
    \item If so in addition to being factually helpful, pair with “helpfulUnbiasedLanguage” or “helpfulEmpathetic” as needed.
    \item If tone, speculation, or bias prevents meaningful clarification, pick the corresponding “notHelpful” reason.
\end{itemize}
4. Is the note hard to understand, incomplete, or not addressing the claim?
\begin{itemize}[leftmargin=*,topsep=2pt,noitemsep]
\item Assign “notHelpfulHardToUnderstand” or “notHelpfulOffTopic” as required.
\end{itemize}
5. Would a typical, reasonably attentive reader gain essential, accurate insight into the claim’s truth, context, or credibility—including debunking of misused/incorrect supporting evidence—because of this note?
\begin{itemize}[leftmargin=*,topsep=2pt,noitemsep]
    \item If yes, “helpful” reasons most fitting the note’s substance.
    \item If no, most directly explanatory “notHelpful” reasons. 
\end{itemize}
Output in the following JSON format only, no extra output:
\{"helpfulness": helpful or non\_helpful,"reasons":"reason1;reason2"\}\newline
CLAIM: \$\{claim\}\newline
NOTE: \$\{note\}\newline                     
Answer: \\
\end{longtable}

% seed definitions
\begin{longtable}{p{1\columnwidth}}
\toprule
\textbf{Seed Definitions} \\
\midrule
\endfirsthead

\toprule
\textbf{Seed Definitions} \\
\midrule
\endhead

\midrule 
\multicolumn{1}{r}{{Continued on next page}} \\
\endfoot

\bottomrule
\\[\abovecaptionskip]
\caption{Detailed seed definitions that explain why notes are helpful and unhelpful.}
\label{tab:seed_def}
\endlastfoot

\textbf{Helpful Reasons}
\begin{itemize}[leftmargin=*,topsep=2pt,noitemsep]
    \item "helpfulClear": "**HelpfulClear** refers to the practice of validating and clarifying claims by providing accurate context or corrections through credible evidence and expert opinion. It involves critically examining information to dispel misinformation, address misconceptions, and ensure that communications are both accurate and informative. The aim is to enhance understanding, promote informed decision-making, and prevent the spread of false or misleading narratives.",
    \item "helpfulGoodSources": "**HelpfulGoodSources** refers to reliable, fact-based information that clarifies, corrects, or disproves misleading or false claims. It encompasses various forms of authoritative references, including government reports, scientific studies, expert opinions, and credible news articles. The purpose of these sources is to provide accuracy and context to statements made in public discourse, aiding in the dissemination of factual knowledge and supporting informed decision-making.",
    \item "helpfulAddressesClaim": "**HelpfulAddressesClaim** is a concept that refers to the clarification or correction of claims made in various contexts by providing accurate information, context, or details that counter or explain the original assertion. This often involves highlighting misconceptions, providing factual evidence, or referencing authoritative sources to support the revised understanding of the claim, ultimately helping to prevent misinformation and promoting informed discourse.",
    \item "helpfulImportantContext": "The REASON, defined as \"helpfulImportantContext,\" refers to the provision of essential background information that clarifies or corrects claims presented in various statements. This context is crucial for understanding the accuracy, validity, or implications of the claims made, by offering factual corrections, historical context, or relevant comparisons to ensure a more informed and nuanced comprehension of the situation or assertion being discussed.",
    \item "helpfulInformative": "The definition of the REASON \"helpfulInformative\" is that it refers to explanations or notes provided in response to claims, which serve to clarify, correct inaccuracies, or provide essential context and background information. These notes help the reader better understand the claims by presenting factual evidence, linking to credible sources, and addressing misunderstandings, thereby enhancing the overall comprehension of the topic at hand.",
    \item "helpfulUnbiasedLanguage": "The definition of the REASON \"helpfulUnbiasedLanguage\" is the use of clear, neutral, and factual language that provides context and clarification to claims without introducing bias or misinformation. It seeks to present information in an objective manner, ensuring that the audience receives accurate and relevant details to better understand the claims being made. This approach emphasizes transparency and factual integrity, allowing for informed discussions and assessments.",
    \item "helpfulEmpathetic": "**Helpful Empathetic**: This reason reflects the intention to provide accurate and clarifying information in response to claims that are misleading or false. It embodies a commitment to enhancing understanding by offering factual corrections and context that reveal the truth, thereby enabling informed discussion. The goal is to support others in recognizing discrepancies or inaccuracies in claims, fostering a more informed perspective while acknowledging the importance of factual accuracy."
\end{itemize} \\
\textbf{Unhelpful Reasons}
\begin{itemize}[leftmargin=*,topsep=2pt,noitemsep]
    \item "notHelpfulArgumentativeOrBiased": "The REASON "notHelpfulArgumentativeOrBiased" can be defined as follows: This classification refers to comments or notes that fail to provide constructive or relevant context to the accompanying claims. Instead, they tend to be argumentative, inflammatory, or biased, often lacking factual support or logical reasoning. Such responses may attempt to undermine the original claim without addressing its merits, focusing instead on personal attacks, irrelevant comparisons, or emotional appeals that detract from a rational discussion. In essence, they do not foster an understanding of the claim and instead perpetuate division or misinformation.",
    \item "notHelpfulMissingKeyPoints": "The definition of the REASON \"notHelpfulMissingKeyPoints\" is: The notes provided do not adequately address or clarify the claims made. Instead, they often introduce unrelated information or misconstrue the context, leaving significant gaps in understanding the primary arguments presented in the claims.",
    \item "notHelpfulNoteNotNeeded": "The definition of the REASON \"notHelpfulNoteNotNeeded\" is that the provided notes fail to contribute meaningful or relevant information to clarify, support, or contradict the claims made. Instead, they either introduce unrelated details, emphasize inaccuracies without addressing the core of the claim, or illustrate a lack of factual backing, thus rendering them ineffective in providing substantiated context or understanding regarding the claims.",
    \item "notHelpfulIncorrect": "The definition of the REASON \"notHelpfulIncorrect\" is as follows: This category refers to claims that are supported by notes or explanations that are factually inaccurate, misleading, or irrelevant to the initial claim. The notes do not effectively clarify, substantiate, or provide accurate context for the claims, thereby failing to contribute meaningful information to the discussion. Instead, they may promote misinformation, misunderstandings, or misinterpretations of the original assertions.",
    \item "notHelpfulSourcesMissingOrUnreliable": "The REASON for the samples being non-helpful is defined as **the presence of sources that are unreliable, lack credibility, or do not substantiate the claims made**. This includes situations where evidence is anecdotal, opinion-based, or referencing non-verifiable documents, thereby failing to provide the necessary backing to support the claims.",
    \item "notHelpfulSpamHarassmentOrAbuse": "The definition of the REASON \"notHelpfulSpamHarassmentOrAbuse\" is content that is either unconstructive, irrelevant, or disrespectful in nature, often resorting to personal attacks, derogatory language, or baseless accusations. Such comments fail to provide meaningful engagement or contribute to the discussion, instead serving to provoke or demean individuals, typically lacking in factual or supportive content.",
    \item "notHelpfulIrrelevantSources": "The REASON \"notHelpfulIrrelevantSources\" can be defined as follows: NotHelpfulIrrelevantSources refers to a category of information where the claims being made are accompanied by explanations, notes, or sources that do not provide pertinent or factual support to validate the claims. This includes instances where the sources cited are either misleading, unrelated to the claim, or factually incorrect, resulting in a lack of credible evidence that would otherwise substantiate the assertions made. Essentially, information categorized under NotHelpfulIrrelevantSources fails to enhance understanding or clarify the claims due to its inaccuracy or irrelevance.",
    \item "notHelpfulOther": "The definition of the REASON \"notHelpfulOther\" is that the provided notes do not effectively contribute to or clarify the claims made. Instead, they often reference unrelated information or distract from the central claim, failing to provide direct support, context, or factual information pertinent to understanding or verifying the claim.", 
    \item "notHelpfulHardToUnderstand": "The REASON for the above samples being non-helpful is that the notes provided do not effectively clarify or support the claims made. They often lack relevant information, context, or clarity, making it difficult for readers to comprehend the claims or understand their significance. This results in a communication gap where the intended message is obscured or misunderstood.",
    \item "notHelpfulOpinionSpeculationOrBias": "The definition of the REASON \"notHelpfulOpinionSpeculationOrBias\" is as follows: This reason applies to statements or claims that are influenced by subjective views, personal opinions, speculation, or perceived biases rather than presenting objective facts or useful information. Such claims often lack substantiation and do not contribute constructively to the discourse, focusing instead on emotional responses or distorted interpretations of events or behaviors."
\end{itemize} \\
\end{longtable}

\begin{longtable}{p{1\columnwidth}}
\toprule
\textbf{Optimized Definitions} \\
\midrule
\endfirsthead

\multicolumn{1}{c}%
{{\bfseries Table \thetable\ continued from previous page}} \\
\toprule
\textbf{Optimized Definitions} \\
\midrule
\endhead

\midrule 
\multicolumn{1}{r}{{Continued on next page}} \\
\endfoot

\bottomrule
\\[\abovecaptionskip]
\caption{Detailed optimized definitions that explain why notes are helpful and unhelpful.} \label{tab:optimized_def}
\endlastfoot

\textbf{Helpful Reasons}

\begin{itemize}[leftmargin=*,topsep=2pt,noitemsep]
    \item helpfulClear: The note delivers an accurate, unambiguous correction or direct clarification targeting a central claim or supporting fact/evidence. Use when the note dispels a major misconception, corrects a key number, identifies a misattributed element, or plainly states the essential fact needed to resolve or accurately interpret the claim.

    \item helpfulAddressesClaim: The note explicitly and accurately responds to or explains the principal assertion, central factual sub-claim, or foundational evidence of the claim. Includes refuting an implied mechanism or performing a core fact check that settles the claim's credibility. 

    \item helpfulGoodSources: The note cites sources that are authoritative, reputable, and faithfully, directly, and unambiguously support its factual correction or clarification. Pick only if sourcing is integral—don't select if other reasons are more fundamental to helpfulness.

    \item helpfulImportantContext: The note supplies indispensable, targeted background or context needed for the reader to assess the claim or its evidence correctly (such as revealing regulatory nuance, historical policies, deception/fraud, or other context that transforms meaning or evaluation).

    \item helpfulInformative: The note gives essential factual explanation or contextualization directly linked to dispelling misunderstanding about the claim's main point. Don't use if the note merely provides general or background info.

    \item helpfulUnbiasedLanguage: The note's language is clear, objective, and neutral while providing a substantive correction or clarification. Use only paired with a direct factual "helpful" label.

    \item helpfulEmpathetic: The note enhances the factual correction by presenting it in a manner that increases reader trust and comprehension through clarity and respect.

    \item helpfulUniqueContext: The note delivers information not widely available, central to correctly interpreting the claim or core evidence.
\end{itemize}

\textbf{Unhelpful Reasons}
\begin{itemize}[leftmargin=*,topsep=2pt,noitemsep]
    \item notHelpfulArgumentativeOrBiased: Hostile, inflammatory, or overtly biased language dominates and no factual clarification or relevant context is given.

    \item notHelpfulMissingKeyPoints: The note wholly fails to clarify, correct, or address the main assertion, supporting fact, or evidence, OR it focuses solely on peripheral, partial, or unrelated points.

    \item notHelpfulNoteNotNeeded: The note adds no meaningful, fresh, or essential clarification—it is trivial, redundant, tangential, or consists of widely-known facts non-essential for comprehending the claim.

    \item notHelpfulIncorrect: Contains factual errors, misleading/unsupported claims, or confuses main assertions or evidence.

    \item notHelpfulSourcesMissingOrUnreliable: Sources are missing, non-credible, don't support the note's claim, or are misrepresented.

    \item notHelpfulSpamHarassmentOrAbuse: Contains spam, personal attacks, or offensive/unrelated content.

    \item notHelpfulIrrelevantSources: Sources in the note do not substantiate or clarify any main aspect of the claim or are not relevant to its core.

    \item notHelpfulOther: Any failure to clarify or support the claim not captured by the above categories.

    \item notHelpfulHardToUnderstand: The note is so unclear, incomplete, or poorly written that its meaning regarding the claim cannot be determined.

    \item notHelpfulOpinionSpeculationOrBias: Personal views, speculation, or bias present to such a degree that no clarification on the main claim or evidence is possible.

    \item notHelpfulOffTopic: The note is wholly unrelated to any claim feature or its main supporting facts/evidence. 
\end{itemize}
\end{longtable}

% Fact Checking Prompt
\begin{longtable}{p{1\columnwidth}}
\toprule
\textbf{Prompt for Fact Checking Climate Fever claims} \\
\midrule
\endfirsthead

\multicolumn{1}{c}%
{{\bfseries Table \thetable\ continued from previous page}} \\
\toprule
\textbf{Prompt for Fact Checking Climate Fever claims} \\
\midrule
\endhead

\midrule 
\multicolumn{1}{r}{{Continued on next page}} \\
\endfoot

\bottomrule
\\[\abovecaptionskip]
\caption{Prompt for fact-checking \climatefever claims.} \label{tab:fc_prompt}
\endlastfoot
\textbf{Direct Prompt}\newline
Fact-check the following claim using provided evidence:\newline
Claim: \$\{claim\}\newline
Evidence: \$\{evidence\_text\}\newline
Classify the claim as \textit{SUPPORTS}, \textit{REFUTES}, \textit{NOT\_ENOUGH\_INFO} or \textit{DISPUTED}.\newline
Format: Classification: [YOUR\_ANSWER]\newline
Brief reason: \newline

\textbf{Incorporating Helpfulness Information}\newline
Fact-check this claim using the evidence and the helpfulness information of the evidence, if the evidence is not helpful, take less weight of the evidence.\newline
Claim: \$\{claim\}\newline
Evidence: \$\{evidence\_text\_with\_helpfulness\_information\}\newline
Classify the claim as \textit{SUPPORTS}, \textit{REFUTES}, \textit{NOT\_ENOUGH\_INFO} or \textit{DISPUTED}.\newline
Format: Classification: [YOUR\_ANSWER]\newline
Brief reason:\newline
\end{longtable}

\end{document}